\documentclass[letterpaper]{article} 
\usepackage{aaai25}  
\usepackage{times}  
\usepackage{helvet}  
\usepackage{courier}  
\usepackage[hyphens]{url}  
\usepackage{graphicx} 
\usepackage{natbib}  
\usepackage{caption} 
\usepackage{algorithm}
\usepackage{algorithmic}

\usepackage{amsfonts}
\usepackage{booktabs}
\usepackage{multirow}
\usepackage{threeparttable}
\usepackage{tcolorbox}
\usepackage{fvextra}
\usepackage{xcolor}
\usepackage{soul} 

\DefineVerbatimEnvironment{VerbatimWrap}{Verbatim}
{
    fontfamily=zi4, 
    fontsize=\small, 
    breaklines=true, 
    breaksymbolleft={},
    commandchars=\\\{\}, 
}

\usepackage{xcolor}

\usepackage{newfloat}
\usepackage{listings}
\DeclareCaptionStyle{ruled}{labelfont=normalfont,labelsep=colon,strut=off} 
\floatstyle{ruled}
\newfloat{listing}{tb}{lst}{}
\floatname{listing}{Listing}
\title{A Multi-Stage Framework with Taxonomy-Guided Reasoning for Occupation Classification Using Large Language Models}

\author{
    Palakorn Achananuparp,\textsuperscript{\rm 1}
    Ee-Peng Lim,\textsuperscript{\rm 1}
    Yao Lu\textsuperscript{\rm 2}
}
\affiliations{
    \textsuperscript{\rm 1}Singapore Management University\\
    \textsuperscript{\rm 2}Columbia University\\
    palakorna@smu.edu.sg, eplim@smu.edu.sg, yl2479@columbia.edu
}

\begin{document}

\maketitle

\begin{abstract}
Automatically annotating job data with standardized occupations from taxonomies, known as occupation classification, is crucial for labor market analysis. However, this task is often hindered by data scarcity and the challenges of manual annotations. While large language models (LLMs) hold promise due to their extensive world knowledge and in-context learning capabilities, their effectiveness depends on their knowledge of occupational taxonomies, which remains unclear. In this study, we assess the ability of LLMs to generate precise taxonomic entities from taxonomy, highlighting their limitations, especially for smaller models. To address these challenges, we propose a multi-stage framework consisting of inference, retrieval, and reranking stages, which integrates taxonomy-guided reasoning examples to enhance performance by aligning outputs with taxonomic knowledge. Evaluations on a large-scale dataset show that our framework not only enhances occupation and skill classification tasks, but also provides a cost-effective alternative to frontier models like GPT-4o, significantly reducing computational costs while maintaining strong performance. This makes it a practical and scalable solution for occupation classification and related tasks across LLMs.   
\end{abstract}

\section{Introduction}
\label{sec:introduction}

The Standard Occupational Classification (SOC) systems and associated taxonomies, such as the O*NET-SOC taxonomy (O*NET-SOC), are essential to labor market analysis and workforce research \cite{hu2016language,guo2024evolution,kalhor2024gender}. These systems provide a systematic framework for categorizing occupations, jobs, skills, and job relevant knowledge. By mapping labor-market data, such as job postings, resumes, or employment surveys, to standardized occupations (referred to as SOC codes and their corresponding SOC titles), researchers, policymakers, and businesses can gain useful insights into labor market trends and expectations. These insights are instrumental for shaping policies, optimizing workforce planning, and guiding industry/manpower development initiatives. 

However, automatic occupation classification is inherently challenging due to the scarcity of large publicly available labeled datasets which require laborious efforts of labor market experts. The lack of data resources further hinders the development and evaluation of robust supervised learning-based classification models. Given the sheer size and complexity of occupational taxonomies, non-expert annotators would require rigorous training to effectively interpret and apply the nuanced structures of these classifications, further complicating the issues.

Recent advancements in large language models (LLMs) have shown promises for addressing these challenges through prompt engineering \cite{li2023llm4jobs,d2024context}. Despite their potential, critical research gaps remain unaddressed. First, the extent to which LLMs rely on their knowledge of occupational taxonomies for classification has not been explored. LLMs are pre-trained on broad datasets, which may include partial representations of occupational taxonomies. The quality and coverage of this taxonomic knowledge can also vary significantly across LLMs due to differences in their training data, architectures, and model sizes. Consequently, their classification effectiveness may be inconsistent. This raises important questions about the reliability and accuracy of LLM-based methods, particularly when dealing with less common or highly specific occupational categories, where the knowledge gaps are likely to be most pronounced.

Moreover, existing LLM-based methods often rely on simplistic prompting techniques, such as direct queries \cite{li2023llm4jobs} or limited in-context examples \cite{d2024context}, which fail to fully utilize the rich information embedded in taxonomies. Taxonomies like O*NET-SOC contain detailed descriptions and hierarchical structures that can enhance classification accuracy. However, current methods fail to integrate this information effectively. While chain-of-thought (CoT) reasoning~\cite{wei2022chain} is popular for general-purpose prompting technique, its adaptation to domain-specific tasks like occupation classification remains unclear. CoT decomposes problems into logical steps but does not inherently exploit taxonomic knowledge, limiting the performance of existing approaches.

To address these research gaps, we conduct a knowledge assessment study to evaluate the capabilities of various LLMs in generating accurate SOC titles and codes from the ONET-SOC taxonomy. Our findings reveal two major shortcomings: (1) LLMs, particularly smaller models, often struggle with generating precise SOC codes, particularly for less common occupations, and (2) their classification accuracy varies significantly based on hierarchical granularity, indicating limited internalization of taxonomic knowledge. Building on these findings, we propose a multi-stage framework that integrates inference, retrieval, and reranking components. Specifically, we incorporate external knowledge from the O*NET taxonomy to guide LLM's reasoning through grounded in-context examples. Unlike computationally intensive fine-tuning or complex model modifications, our framework relies on efficient prompting and retrieval strategies, making it adaptable to a variety of LLMs without requiring additional training. By leveraging the detailed class information embedded in ONET-SOC, this framework enhances the accuracy and robustness of occupation classification tasks. Finally, we evaluate the framework on both occupation and skill classification tasks, demonstrating its effectiveness in mitigating the knowledge gaps and its adaptability across domains.

Our contributions are fourfold:
\begin{enumerate}
    \item We are the first to systematically investigate LLM knowledge of the O*NET-SOC taxonomy through a series of generation tasks, providing insights into their strengths and limitations.
    \item We propose a novel framework that builds on retrieval-based prompting methods \cite{li2023llm4jobs,d2024context} by systematically integrating grounded rationals into in-context examples, aligning LLM reasoning and outputs with taxonomic information while minimizing prompt engineering effort.
    \item We explicitly examine the relationship between LLMs' taxonomic knowledge and their downstream classification performance. Our findings demonstrate that leveraging external taxonomic information throughout the classification pipeline, from inference to reranking, enhances alignment with the taxonomy and improves overall performance across different LLMs.
    \item We demonstrate that our framework significantly reduces the computational costs while maintaining high classification accuracy. By enabling smaller, more efficient LLMs to achieve performance comparable to frontier models like GPT-4o, the framework provides a scalable and cost-effective alternative to GPT-4o's direct prompting, making it practical for large-scale classification tasks.
\end{enumerate}

\section{Related Work}
\label{sec:related_work}

\subsection{Assessing LLM Knowledge}

Existing research has investigated the ability of LLMs to memorize factual knowledge in the training data mainly through learning from high-frequency patterns, and their struggle with low-resource tasks \cite{tanzer2022memorisation}. Researchers have developed methods to systematically evaluate LLM knowledge of both common and long-tail facts in generic and specialized domains by creating tasks derived from knowledge graphs. \cite{sun2023head,luo2023systematic,kumar2024automatic}. In the context of taxonomic knowledge, recent work has examined LLMs' understanding of hierarchical structures in specialized domains. For example, Sun et al. \cite{sun2024large} developed binary and multiple-choice questions to comprehensively test frontier LLMs' comprehension of entities across ten taxonomies, spanning domains such as shopping, language, and medicine. Similarly, Soroush et al. \cite{soroush2024large} assessed frontier LLMs' capabilities in generating structured codes from medical taxonomies. Both studies revealed limitations in frontier LLMs's ability to generate taxonomic entities from specialized domains.

Our study is the first to examine LLM knowledge of occupational taxonomies. Our task design is more comprehensive than previous work, encompassing those of Sun et al. \cite{sun2024large} and Soroush et al. \cite{soroush2024large}. Additionally, we introduce a task variant that tests LLM performance when partial hints are provided. Moreover, our investigation focuses on efficient LLMs, which are practical for large-scale downstream applications such as our occupation classification.

\subsection{Occupation and Skill Classifications}
\label{subsec:related_work_classification}

Occupation classification has been explored in numerous studies using proprietary datasets and various methodologies to map job-related data to standardized occupations. Traditional approaches often involve multi-class text classification that rely heavily on feature engineering techniques, such as bag-of-words and word embeddings \cite{boselli2017using}, combined with standard machine learning models like SVM and random forest \cite{russ2016computer,javed2015carotene}. Some studies focus on classifying occupations based on job titles \cite{boselli2017using,javed2015carotene}, while others use richer input data, such as full job postings \cite{li2023llm4jobs}. 

Recent work has integrated LLMs in the occupation classification pipeline. For example, LLM4Jobs \cite{li2023llm4jobs} employed a two-stage framework, utilizing Vicuna 33B to infer occupational labels from job postings and retrieve top-ranked candidates from the ESCO taxonomy. Similarly, LLMs have been applied to workforce-related tasks like multi-label skill classification, including generating synthetic training data for fine-tuning \cite{clavie2023large,decorte2023extreme}, inferring standardized skill labels \cite{d2024context}, and reranking candidates \cite{clavie2023large,d2024context}.

To our knowledge, this study is one of the few to evaluate occupation classification on large-scale job records derived from diverse, real-world online resume data, involving fine-grained SOC labels from the O*NET-SOC taxonomy. Unlike fine-tuned systems \cite{zhang2022kompetencer,clavie2023large,decorte2022design}, our framework is completely training-free and most comparable to LLM4Jobs \cite{li2023llm4jobs} and Infer-Retrieve-Rank (IReRa) \cite{d2024context}, where LLMs are utilized in conjunction with a taxonomy, along with retrieval and reranking modules in a classification pipeline. However, it differs from previous work in a few key aspects. 

Firstly, our framework uniquely focuses on aligining LLM reasoning with taxonomic knowledge through taxonomy-guided in-context examples, which significantly enhances the overall accuracy. In contrast, LLM4Jobs and IReRa simply utilize zero-shot and general-purpose CoT prompting approaches, respectively. Secondly, we leverage rich exterrnal taxonomic information in the retrieval stage, a strategy similar to LLM4Jobs but not utilized in IReRa. This overcomes the limited taxonomic knowledge of many existing LLMs.  Lastly, while LLM4Jobs focuses solely on retrieval after inference, our framework, like IReRa, introduces an additional reranking stage, where LLMs are used to further refine the list of retrieved candidates.

\section{Knowledge Assessment Study}
\label{sec:knowledge_assessing}

We begin by systematically examining the extent to which LLMs memorize and internalize occupational taxonomies through a series of generation tasks. For this investigation, we chose the O*NET-SOC taxonomy -- a widely adopted hierarchical classification system for standardized occupations -- and evaluated seven LLMs, which are described in detail in subsequent sections.

\subsection{The O*NET-SOC Taxonomy}
\label{subsec:onet_taxonomy}
The O*NET-SOC taxonomy (O*NET-SOC) is a hierarchical system for organizing occupations within the Standard Occupational Classification (SOC) framework. Each occupation is represented by a unique SOC code, a title, a detailed description, and other associated information. The taxonomy comprises five levels of granularity, all encoded in the SOC codes: \textbf{major groups} (2-digit codes), \textbf{minor groups} (3-digit codes), \textbf{broad occupations} (5-digit codes), \textbf{detailed occupations} (6-digit codes), and \textbf{specific occupations} (8-digit codes). Table ~\ref{tab:soc_occupation} illustrates its hierarchical structure using Data Scientists as an example. This study focuses on \textbf{O*NET-SOC 2019}, the most recent version, which includes 1,016 specific occupations.

\begin{table}[ht]
\caption{Hierarchical breakdown of the SOC code for the Data Scientists occupation (15-2051.00)}
\centering
\begin{tabular}{@{}lllll@{}}
\toprule
SOC Code & Major & Minor & Broad & Detailed \\ \midrule
15-2051.00 & 15 & 15-2 & 15-205 & 15-2051 \\ \bottomrule
\end{tabular}
\label{tab:soc_occupation}
\end{table}

\subsection{Choices of LLMs}
\label{subsec:llm}

We selected seven LLMs, comprising two frontier models: \textbf{GPT-4o} and \textbf{Gemini-1.5 Pro}, and five efficient models: \textbf{GPT-3.5 Turbo}, \textbf{Gemini-1.5 Flash}, \textbf{Claude Instant 1.2}, \textbf{Llama 3 8B Instruct}, and \textbf{Mistral Small}. 

The frontier models were included to establish an upper bound on performance. These models represent the state-of-the-art in LLMs at the time of this study and are characterized by their significantly larger parameter sizes and more advanced architectures. In contrast, the efficient models were chosen for their balance between performance and computational cost. These models tend to have smaller parameter sizes or optimized architectures, enhancing their token processing speed and reducing inference costs. Their efficiency makes them particularly practical for large-scale downstream classification tasks such as occupation classification.

\subsection{Task Design}
\label{subsec:task_design}
We devise a set of generation tasks to assess LLM knowledge by instructing LLMs to generate SOC titles and codes in order to assess both the ability to directly retrieve taxonomy information and  the capacity to perform at varying levels of granularity within the taxonomy.

First, we structure the tasks into two modes of generation: \textbf{recall} and \textbf{recognition}. Recall tasks evaluate the model's ability to freely generate titles or codes, while recognition tasks assess its ability in selecting the correct output from a list of candidates. Recall tasks are further divided into \textbf{complete} and \textbf{partial} recall modes. In complete recall, the model generates the full code or title without additional hints. In partial recall, the model is tasked with completing a partially provided code or title.

Additionally, we focus on two main \textit{context}-to-\textit{output} generation directions: code-to-title (\textbf{CT}) and title-to-code (\textbf{TC}). In CT, we test the model's ability to generate a title (output) from a given code (context), while in the TC direction, we test its ability to generate a code (output) from a given title (context). To analyze model performance across varying levels of granularity, each task direction is examined using both \textbf{8-digit} and \textbf{2-digit} codes. These combinations yield three task variants: \textbf{CT8} (title generation from 8-digit codes), \textbf{TC8} (8-digit code generation from titles), and \textbf{TC2} (2-digit code generation from titles). We exclude CT2 (title generation from 2-digit codes) as multiple titles can correspond to one 2-digit code.

The final task set, displayed in Table ~\ref{tab:knowledge_tasks}, includes CT8, TC8, and TC2 in the complete recall and recognition tasks and CT8 and TC8 in the partial recall tasks. TC2 is excluded from partial recall tasks as a single-digit hint provides insufficient context for meaningful evaluation.

\begin{table}[!th]
\caption{Knowledge Assessment Tasks}
\label{tab:knowledge_tasks}
\centering
\resizebox{\columnwidth}{!}{
\begin{tabular}{@{}lllll@{}}
\toprule
\multirow{2}{*}{Mode} & \multicolumn{2}{c}{Direction} & \multirow{2}{*}{Granularity} & \multirow{2}{*}{Task} \\ \cmidrule(lr){2-3}
 & Context & Output &  &  \\ \midrule
\multirow{3}{*}{Complete Recall} & Code & Title & 8-digit & CT8 \\
\cmidrule(lr){2-5}
 & \multirow{2}{*}{Title} & \multirow{2}{*}{Code} & 8-digit & TC8 \\
 &  &  & 2-digit & TC2 \\ \midrule
\multirow{2}{*}{Partial Recall} & Code & Title & 8-digit & CT8 \\
 & Title & Code & 8-digit & TC8 \\ \midrule
\multirow{3}{*}{Recognition} & Code & Title & 8-digit & CT8 \\
\cmidrule(lr){2-5}
 & \multirow{2}{*}{Title} & \multirow{2}{*}{Code} & 8-digit & TC8 \\
 &  &  & 2-digit & TC2 \\ \bottomrule
\end{tabular}}
\end{table}

\subsection{Research Questions}
\label{subsec:assessing_questions}

Given these configurations, we pose the following research questions:

\textbf{RQ1}: How do frontier and efficient LLMs compare in their ability to generate occupational titles and codes from the O*NET-SOC taxonomy?

\textbf{RQ2}: How does LLM performance vary across different levels of granularity within the O*NET-SOC taxonomy?

\textbf{RQ3}: How do different task design elements, such as generation mode, direction, etc., influence LLM performance in generating titles and codes?

\subsection{Task Generation}
\label{subsec:task_generation}

For each occupation in the taxonomy, we generate recall and recognition tasks using prompt templates, shown in Figure ~\ref{prompt:knowledge} in the appendix. All tasks include a one-shot example to help models follow instructions accurately, which is especially helpful for smaller models.

For partial recall tasks, we provide the first half of the answer as a hint. For example, for the occupation \textit{Data Scientists} (15-2051.00), the prompts for a TC8 partial recall task are: ``\textit{In the O*NET-SOC 2019 taxonomy, the 8-digit code for the title Data Scientists is $<$blank$>$. Hint: It starts with 15-20}.'' 

For recognition tasks, we generate five randomly ordered options: the correct answer corresponding to the given occupation, two candidates randomly sampled from related occupations within the same major (2-digit code) groups, and two candidates randomly sampled from different major groups. 

All experiments were conducted in October 2024. The models from OpenAI (GPT-4o and GPT-3.5 Turbo), Vertex AI (Gemini-1.5 Pro and Gemini-1.5 Flash), and Amazon Bedrock (Claude Instant 1.2, Llama 3 8B Instruct, and Mistral Small), were accessed via their respective APIs. The temperature was set to zero for all tasks and LLMs to ensure deterministic outputs.

\subsection{Metrics and Statistical Analysis}
We evaluate performance using accuracy, defined as the proportion of generated answers that exactly match the corresponding values in O*NET-SOC. The average accuracy score provides an overall metric measuring the model's ability to generate taxonomic entities across tasks. A one-way ANOVA and linear regression were performed using Python's SciPy (1.11.3) and Statsmodels (0.14) libraries. The data used in the analyses satisfy normality assumption.

\subsection{Knowledge Assessment Results}
\label{subsec:assessing_results}

\begin{table*}[ht!]
\caption{Performance of frontier and efficient LLMs on knowledge recall and recognition task variants. The best results are highlighted in bold and the second-best results are underlined.}
\label{tab:assessing_performance_all}
\centering
\begin{tabular}{@{}lrrrrrrrrr@{}}
\toprule
\multicolumn{1}{l}{\multirow{2}{*}{LLM}} & \multicolumn{3}{c}{Complete Recall} & \multicolumn{2}{c}{Partial Recall} & \multicolumn{3}{c}{Recognition} & \multicolumn{1}{c}{\multirow{2}{*}{Average}} \\ \cmidrule(lr){2-9}
\multicolumn{1}{l}{} & \multicolumn{1}{c}{CT8} & \multicolumn{1}{c}{TC8} & \multicolumn{1}{c}{TC2} & \multicolumn{1}{c}{CT8} & \multicolumn{1}{c}{TC8} & \multicolumn{1}{c}{CT8} & \multicolumn{1}{c}{TC8} & \multicolumn{1}{c}{TC2} & \multicolumn{1}{c}{} \\ \midrule
GPT-4o & \textbf{0.8612} & \textbf{0.9104} & \textbf{0.9754} & \textbf{0.9626} & \textbf{0.873} & \textbf{0.9242} & \textbf{0.939} & 0.9232 & \textbf{0.9211}\\
Gemini-1.5 Pro & 0.2923 & 0.5453 & 0.8858 & \underline{0.8573} & 0.6083 & 0.7067 & \underline{0.8907} & \textbf{0.9685} & 0.7194 \\ \midrule
GPT-3.5 Turbo & \underline{0.5551} & \underline{0.6742} & \underline{0.9183} & 0.8041 & \underline{0.6693} & 0.3878 & 0.8888 & \underline{0.9587} & \underline{0.7320} \\
Gemini-1.5 Flash & 0.002 & 0.1004 & 0.7333 & 0.5522 & 0.1614 & 0.2008 & 0.3622 & 0.9439 & 0.3820 \\
Claude Instant 1.2 & 0.1663 & 0.3809 & 0.6663 & 0.6378 & 0.3661 & 0.373 & 0.6831 & 0.8829 & 0.5196 \\
Llama 3 8B Instruct & 0.0118 & 0.0768 & 0.247 & 0.3022 & 0.1919 & 0.3819 & 0.5846 & 0.439 & 0.2794 \\
Mistral Small & 0.0039 & 0.2185 & 0.7441 & 0.4734 & 0.248 & 0.1437 & 0.4813 & 0.8524 & 0.3957 \\ \bottomrule
\end{tabular}
\end{table*}

Table ~\ref{tab:assessing_performance_all} presents the performance of the seven LLMs on the complete recall, partial recall, and recognition tasks. Overall, GPT-4o is the top performer, achieving an impressive average accuracy of 0.9211. Notably, GPT-3.5 Turbo follows as the second-best model with an average accuracy of 0.732, outperforming a more advanced model like Gemini-1.5 Pro. In contrast, Llama 3 8B Instruct has the lowest performance with an average accuracy of 0.2794. Most models perform better in recognition than recall tasks. Moreover, complete recall are generally more challenging than partial recall tasks.

\textbf{RQ1: How do frontier and efficient LLMs compare in their ability to generate occupational titles and codes from the O*NET-SOC taxonomy?}

A one-way ANOVA was conducted to examine differences between frontier and efficient models. The results indicate a statistically significant difference between the performance scores of the frontier and efficient models ($F = 20.9$, $p < 0.001$). Specifically, the frontier models consistently achieve superior performance across all recall and recognition tasks, with GPT-4o setting a significant upper bound. The findings highlight the extensive knowledge and capabilities of frontier models in understanding and accurately generating titles and codes from the O*NET-SOC taxonomy.

\textbf{RQ2: How does LLM performance vary across different levels of granularity within the O*NET-SOC taxonomy?}

To analyze how LLM performance varies across different levels of granularity, a one-way ANOVA was performed to compare LLM accuracy on tasks involving 8-digit (specific occupations) and 2-digit (major occupational groups) SOC codes. The results demonstrate a statistically significant difference between the two levels ($F = 12.667$, $p < 0.001$), with LLMs performing significantly better on 2-digit tasks. The findings suggest that LLM performance varies with the granularity of the occupational codes. This disparity may be influenced by several factors, including the inherent difficulty of predicting finer-grained categories and potential limitations in the LLMs' ability to fully leverage hierarchical information within the taxonomy. 

\textbf{RQ3: How do different task design elements, such as generation mode, direction, etc., influence LLM performance in generating titles and codes?}

To explore how task design influences LLM performance in generating occupational titles and codes, we fit two Ordinary Least Squares (OLS) regression models with accuracy scores as the dependent variable.

 \textbf{Recall vs. recognition}: In the first model, the independent variables included individual LLM models (e.g., GPT-4o), generation mode (recall or recognition), output type (title or code), code granularity (8-digit or 2-digit), and interaction terms to capture combined effects. An N-1 dummy coding was applied for categorical variables, with \textit{Llama 3 8B Instruct}, \textit{recall}, \textit{code}, and \textit{2-digit} as the reference categories for LLM, generation mode, output type, and code granularity, respectively. The model explained a significant proportion of variance in accuracy scores ($N = 52,R^2 = 0.759$).

Adjusting for other factors, significant main effects were found for LLM model, code granularity, and the generation mode and output type interaction. Firstly, the performance of individual LLMs showed significant differences. GPT-4o ($\beta = 0.642$, $p < 0.001$), Gemini-1.5 Pro ($\beta = 0.44$, $p < 0.001$), GPT-3.5 Turbo ($\beta = 0.453$, $p < 0.001$), and Claude Instant 1.2 ($\beta = 0.24$, $p < 0.01$) outperformed the baseline Llama 3 8B Instruct, whereas Gemini-1.5 Flash and Mistral Small did not show significant improvement over the baseline. This finding aligns with the results of RQ1, which demonstrated that frontier models significantly outperform efficient models across tasks.

Secondly, code granularity significantly affects accuracy. Tasks involving 8-digit codes were more challenging ($\beta = -0.308$, $p < 0.001$) compared to 2-digit codes.  The finding is consistent with previous analyses in RQ2, confirming that tasks requiring greater taxonomic specificity are significantly more challenging.

Lastly, there is a significant interaction between recognition mode and title generation ($\beta = -0.277$, $p < 0.05$), suggesting that recognizing valid titles is more challenging than recognizing codes. This disparity is likely due to titles tend to require nuanced linguistic understanding, whereas codes are more structured and deterministic, making them easier for LLMs to match accurately.

\textbf{Complete vs. partial recalls}: To examine the influence of recall modes, a second OLS regression model was fit using the same variables but replacing generation mode with recall mode (\textit{complete recall} as the reference category). 
The analysis ($N=35,R^2 = 0.921$) yielded consistent main effects for LLM model and code granularity. Additionally, significant effects were found for output type and the recall mode and output type interaction. Specifically, titles were more challenging to generate than codes in recall tasks ($\beta = -0.072$, $p < 0.05$). While there was no overall difference in accuracy between complete and partial recall, partial recall significantly improved performance for title generation compared to complete recall ($\beta = 0.355$, $p < 0.001$).

Altogether, these findings suggest that while current LLMs exhibit varying degrees of knowledge regarding the O*NET-SOC taxonomy, their performance in tasks involving the taxonomy varies, with GPT-4o demonstrating the strongest performance. Factors such as task design elements, particularly code granularity and output type, should be considered when leveraging LLMs for downstream applications like occupation classification.

\section{Framework}
\label{sec:framework}

\begin{figure*}[h]
    \centering
    \includegraphics[width=0.6\textwidth]{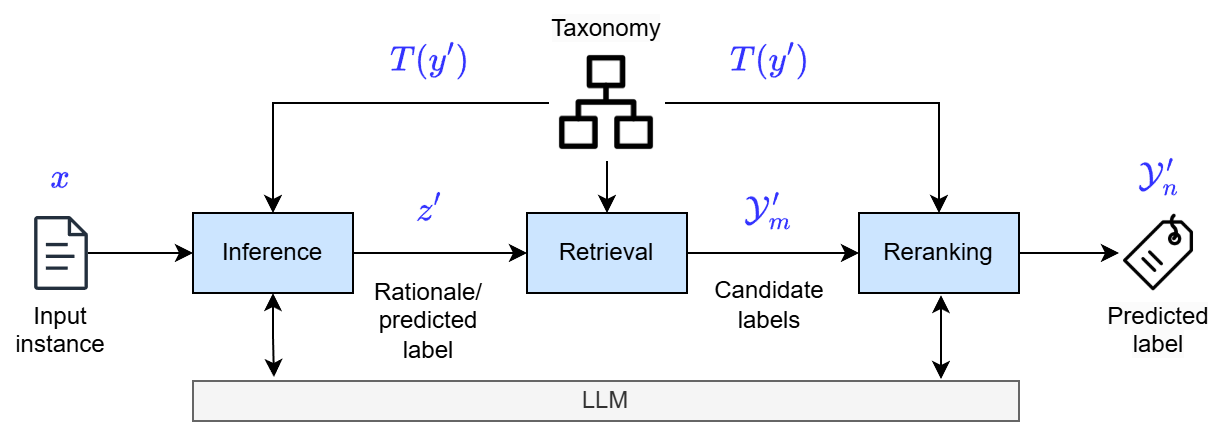}
    \caption{The proposed framework}
    \label{fig:framework}
\end{figure*}

We propose a multi-stage framework for occupation classification, consisting of \textbf{inference}, \textbf{retrieval}, and \textbf{reranking} stages, shown in Figure~\ref{fig:framework}. Let $\mathbb{X}$ denote the input space (e.g., job title-company pairs), $\mathbb{Y}$ the label space (e.g., SOC titles and codes), and $\mathbb{T}$ the taxonomy (e.g., the O*NET-SOC taxonomy) from which $\mathbb{Y}$ are drawn. The framework maps an input $x \in \mathbb{X}$ to corresponding labels $y \in \mathbb{Y}$, with $\mathbb{Y} \subseteq \mathbb{T}$.

The framework is specifically designed to address the limited domain-specific knowledge of LLMs by following these guiding principles: (i) leveraging the reasoning capabilities of LLMs to enhance their ability to infer and select correct class labels; and (ii) utilizing detailed class information from the taxonomy, such as class descriptions, to guide the reasoning process and enhance the retrieval of relevant candidate labels. 

To align LLM reasoning, we introduce \textbf{taxonomy-guided reasoning example} (\textbf{TGRE}), a simple method that incorporates \textit{grounded rationales} into in-context examples with minimal prompt engineering. A grounded rationale is an explanation explicitly linked to the taxonomy, clarifying the relationship between the input and the corresponding label. In a typical classification task, an in-context example $(x, y)$ consists of the task input $x$ (e.g., a job title-company pair) and the corresponding ground-truth label $y$ (e.g., an SOC title and code). Our method enriches these examples by injecting a grounded rationale $r$ that explains the connection between $x$ and $y$. We refer to this grounded example $(x,r,y)$ as TGRE. 

To construct TGRE, we define $T(y)$ as a function that retrieves relevant information from the taxonomy corresponding to the ground-truth label $y$. Formally, $T(y) = \{t_1, t_2,...,t_p\}$ where $\{t_1, t_2,...,t_p\}$ represents the set of $p$ taxonomic details associated with $y$, such as class descriptions, class instances, etc. In this study, $T(y)$ returns class descriptions. The retrieved information $T(y)$ is then used to create the grounded rationale $r = f(T(y))$, where $f$ is a formatting function that translates $T(y)$ into a natural language rationale. In this work, we adopt a simple approach where $f(T(y))$ is the verbatim inclusion of the retrieved value $T(y)$. However, more sophisticated formatting functions $f$, such as those that rewrite or summarize taxonomic information, can be used to adapt $T(y)$. For each i-th example $x_i$, we create a grounded example $(x_i,r_i,y_i)$.


\textbf{Inference}: In the first stage, the input instance $x \in \mathbb{X}$ is provided to the LLM along with a domain-specific prompt to infer a candidate class or label. The goal of the inference stage is to infer a candidate label $y' \in \mathbb{Y}$ based on $x$. In addition, the prompt includes \textit{k}-shot TGRE-based in-context examples $\{(x_1, r_1, y_1),...,(x_k, r_k, y_k)\}$ to demonstrate the task and ensure the reasoning aligns closely with the taxonomy. Various methods, such as random selection, can be used to choose the examples from $\mathbb{X}$. The output of the LLM includes both a rationale $r'$ explaining the inference and the inferred class label $y'$. For occupation classification, $x$ is a job title-company pair, $r$ is derived from the occupation description in O*NET-SOC, and $y'$ consists of SOC title and 8-digit code.

\textbf{Retrieval}: From the output generated in the inference stage, we construct a query denoted as $z'$  to retrieve \textit{m} most similar candidates from an external taxonomy $\mathbb{T}$. Here, $z'$ may represent $y'$ (the inferred label), $r'$ (the rationale), or $(r', y')$.  $x$ is not used in the query because it does not exist in the taxonomy $\mathbb{T}$.  Let $E(.)$ denote an embedding function.  To select the candidates from $\mathbb{T}$, we compute the cosine similarity between the embeddings of the query $E(z')$ and taxonomic entities $E(y_i)$ where $\{y_1,y_2,...,y_{|\mathbb{T}|}\} \in \mathbb{T}$. The taxonomic entities are ranked based on their cosine similarity scores, from highest to lowest, and the top-$m$ candidates $\mathcal{Y}'_m$ are selected. Note that $y'$ may already be in $\mathcal{Y}'_m$. For occupation classification, $\mathcal{Y}'_m$ are SOC occupations, represented by titles and 8-digit codes, from O*NET-SOC.

\textbf{Reranking}: In the final stage, the LLM refines the retrieval output by ranking up to \textit{n} candidates ($n<m$) that are most relevant to the input $x$, denoted as $\mathcal{Y}'_n$ ($\mathcal{Y}'_n \subseteq \mathcal{Y}'_m$). The LLM is prompted to rerank the retrieved candidates based on their contextual fit with $x$. The \textit{k}-shot grounded in-context examples, similar to those used in the inference stage, are included in this prompt to demonstrate the task and ensure accurate alignment with the taxonomy. For occupation classification, this involves reranking the retrieved SOC titles and 8-digit codes to determine the most appropriate match for the job title-company pair.

\textbf{Key differences from IReRa}: While both IReRa \cite{d2024context} and our framework employ a three-stage pipeline (inference, retrieval, and reranking), we introduce the following key improvements. Unlike IReRa, which uses CoT reasoning, our framework employs TGRE to directly align rationales with taxonomic knowledge. Moreover, our retrieval stage uses rationale-augmented queries ($z' = (r',y')$) incorporating both inferred labels and grounded rationales, whereas IReRa relies solely on label-based retrieval ($z'=y'$). Finally, IReRa is a general framework for extreme multi-label classification, utilizing the DSPy programming model \cite{khattab2023dspy}. It features an iterative prompt optimization module aimed at enhancing accuracy but at the cost of significantly higher LLM calls and token usage. In contrast, our framework prioritizes efficiency by randomly selecting in-context examples, thereby minimizing computational overhead. 

\section{Occupation Classification Evaluation}
\label{sec:occupation_evaluation}

To evaluate the effectiveness of the proposed framework, we conducted a multi-class occupation classification task. Specifically, the task involves assigning one of the predefined occupations in O*NET-SOC to an input job title-company pair. This task is inherently challenging due to the ambiguity and diversity of job titles and the size and structure of the taxonomy.

\subsection{Dataset}
\label{subsec:occupation_dataset}

We utilized US-based anonymized resume data gathered from various online professional networking platforms such as LinkedIn and Stack Overflow\footnote{https://kb.lightcast.io/en/articles/7153977-global-data-101} by Lightcast, a labor market analytics company. From this resume data, we created a test set, called \textbf{Jobs12K}, by randomly sampling 11,920 job records. Each record consisted of a job title and its corresponding company name. Moreover, another 100 job records were randomly chosen as a validation set for selecting in-context examples. To ensure data quality, we excluded records that met the following criteria: (i) job titles containing non-alphanumeric or non-ASCII characters, and (ii) job titles exceeding seven words in length. After filtering, the average word counts for job records in the test set and validate set were 5.9 words (S.D. = 2.3) and 5.8 words (S.D. = 2.3), respectively. A detailed description of the Lightcast dataset, used for deriving Jobs12K, is included in Section~\ref{sec:lightcast} of the appendix.

\subsection{Experimental Setup}
\label{subsec:occupation_setup}

We tested various combinations of LLMs, reasoning methods, retrieval units, and ablated variants to evaluate the proposed framework.

\textbf{LLMs}: Each method was tested on five efficient LLMs: GPT-3.5 Turbo, Gemini-1.5 Flash, Claude Instant 1.2 , Llama 3 8B Instruct, and Mistral Small. As previously mentioned, these LLMs were chosen for their cost effectiveness, which make them practical for large-scale classification tasks. GPT-3.5 Turbo was accessed in June 2024, while the remaining models were accessed in October 2024. The temperature was set to zero for all tasks and LLMs to reduce randomness and promote deterministic outputs.

\textbf{Reasoning methods}: We include two reasoning methods, \textbf{TGRE} and \textbf{CoT}, in both the inference and reranking stages. Since the Jobs12K dataset is unlabeled, we manually annotated a small sample ($N < 10$) from its validation set to be used in constructing in-context examples. TGRE utilizes TGRE-based in-context examples, augmenting rationales with occupation descriptions from O*NET-SOC. In contrast, CoT uses the chain-of-thought reasoning to generate rationales in the in-context examples without the taxonomy, serving as a competitive baseline. For ablated variants, indicated by single and double dashes, we exclude the reranking module (\textbf{TGRE-} and \textbf{CoT-}) or both the retrieval and reranking modules (\textbf{TGRE-\,-} and \textbf{CoT-\,-}).

\textbf{Retrieval units}: Two retrieval units -- \textbf{sentence} and \textbf{label} -- are included. For the sentence-based retrieval unit, queries $z'$ are rationales and inferred SOC titles and codes ($r' \cup y'$) generated during the inference stage while documents $y_i$ are SOC titles, 8-digit codes, and descriptions from O*NET-SOC. For the label-based retrieval unit, queries $z'$ are inferred titles and codes $y'$ generated during the inference stage and documents $y_i$ are SOC titles and codes. Sentence transformer models \textit{multi-qa-mpnet-base-dot-v1} and \textit{all-mpnet-base-v2} are used for generating embeddings for the sentence and label units, respectively. These models have been shown in our internal validation to produce optimal embeddings for the respective units. In the retrieval and reranking stages, we set both the number of top-$m$ retrieved candidates and top-\textit{n} selected candidates to 10.

\textbf{Other baselines}: Several method variants in the experiment represent approaches used in existing LLM-based classification methods. These include:

\begin{itemize}
    \item \textbf{IReRa}: Our CoT with label-based retrieval variant is equivalent to IReRa \cite{d2024context} where chain-of-thought is utilized to infer class labels and rerank candidates. The  top candidates is retrieved by computing the similarity between the inferred labels and the taxonomic class labels.
    \item \textbf{LLM4Jobs}: Our ablated sentence-based variants that omit the reranking module (e.g., TGRE- and CoT-) are generally comparable to LLM4Jobs \cite{li2023llm4jobs}. Unlike LLM4Jobs, which employs zero-shot prompting, our approach leverages in-context examples. The best ablated variant in the experiment is used to represent LLM4Jobs.
\end{itemize}

Moreover, we compare precision of the default SOC labels in Jobs12K provided by Lightcast (\textbf{LC}). These labels were inferred using proprietary algorithms.

\textbf{In-context examples}: For all variants, we include a one-shot in-context example. For TGRE, we randomly selected an input instance and its associated ground-truth SOC title and code from the validation set. Then, we retrieved the occupation description corresponding to the SOC title and code and included it into the in-context example as the grounded rationale. For CoT, we followed the protocol used in IReRa \cite{d2024context} to ensure a fair comparison by employing their optimization module to select the best $k$-shot in-context examples from the validation set (where $k$ was empirically set to be 1). Then, we used the LLM to generate CoT rationales for the in-context examples. 

All prompt templates and in-context examples are included in Section~\ref{subsec:occupation_prompt_templates} of the appendix.

\subsection{Evaluation Metric}
\label{subsec:occupation_metric}

We used \textbf{Precision@}\textit{K} ($K = 1$), defined as the proportion of correctly classified labels among the top-\textit{K} predicted labels (SOC titles and 8-digit codes). $K=1$ is appropriate as occupation classification typically considers only the top prediction actionable. As $K$=1 and each instance has a single ground-truth label, Precision@1 is equivalent to Recall@1, F1@1, and Accuracy@1. 

Since ground-truth labels for the job records in Jobs12K were not available, we employed \textbf{GPT-4o}, which has been shown to be highly knowledgeable of the O*NET-SOC taxonomy in Section~\ref{sec:knowledge_assessing}, to identify the most relevant label. Specifically, we prompted GPT-4o to select the most appropriate SOC label from the pool of predicted labels generated by all tested methods for each job record. To mitigate positional bias, the lists of candidate labels were shuffled. GPT-4o was also instructed to explicitly indicate if none of the candidates were applicable. This evaluation was repeated for 3 runs, and majority voting was applied to select the final labels. In the case of ties, the final selection was made randomly. The prompt template for occupation selection is included in Figure~\ref{prompt:zeroshot_annotate_occupation} in Section~\ref{subsec:occupation_prompt_templates}.

Consistent with previous findings \cite{liu2023g-eval,sottana2023evaluation}, our internal validation demonstrated that GPT-4o is a reliable evaluator for this task. Specifically, we conducted a comparison between human judgments and GPT-4o's assessments on a small random sample of 100 job records. The human annotator, who is fairly knowledgeable of the O*NET-SOC taxonomy, provided independent annotations, and the results showed 82\% correlation between human and GPT-4o judgments. The strong agreement provides preliminary evidence that GPT-4o can serve as a reliable proxy for human evaluation in this task.

To compare results, one-way ANOVA with Tukey HSD post-hoc analyses were performed using Python's SciPy (1.11.3) and Statsmodels (0.14) libraries. The data used in the analyses are normally distributed.

\subsection{Results}
\label{subsec:occupation_results}

\begin{table*}[!ht]
\caption{Precision@1 for occupation classification. The best results are highlighted in bold. The columns abbreviate the full names of the LLM models: GPT refers to GPT-3.5 Turbo, Gemini to Gemini-1.5 Flash, Claude to Claude Instant 1.2, Llama to Llama 3 8B Instruct, and Mistral to Mistral Small.}
\label{tab:occupation_results}
\centering
\begin{tabular}{@{}lllrrrrr@{}}
\toprule
Baseline & Reasoning & Retrieval & \multicolumn{1}{l}{GPT} & \multicolumn{1}{l}{Gemini} & \multicolumn{1}{l}{Claude} & \multicolumn{1}{l}{Llama} & \multicolumn{1}{l}{Mistral} \\ 
Equivalence & Method & Unit &  & & & & \\ \midrule
 & TGRE & Sentence & \textbf{0.8114} & \textbf{0.7868} & \textbf{0.7616} & \textbf{0.7612} & \textbf{0.7687} \\
 & CoT & Sentence & 0.7685 & 0.6827 & 0.6944 & 0.6593 & 0.6506 \\
\midrule
 & TGRE & Label & 0.7694 & 0.6909 & 0.6841 & 0.5366 & 0.7202 \\
IReRa & CoT & Label & 0.7049 & 0.6618 & 0.6592 & 0.5491 & 0.5068 \\
\midrule
LLM4Jobs & TGRE- & Sentence & 0.6644 & 0.6018 & 0.6084 & 0.5993 & 0.6466 \\
LLM4Jobs & CoT- & Sentence & 0.6384 & 0.508 & 0.4964 & 0.5261 & 0.3918 \\
 & TGRE- & Label & 0.7015 & 0.5238 & 0.581 & 0.4969 & 0.5985 \\
 & CoT- & Label & 0.5658 & 0.5005 & 0.403 & 0.3006 & 0.1356 \\
& TGRE-\,- & None & 0.4875 & 0.0715 & 0.277 & 0.0597 & 0.0407 \\
& CoT-\,- & None & 0.5002 & 0.0684 & 0.2886 & 0.0925 & 0.0857 \\ \bottomrule
\end{tabular}
\end{table*}

The results of occupation classification across methods and models are shown in Table~\ref{tab:occupation_results}. The best performance is achieved by \textbf{TGRE with sentence-based retrieval} on \textbf{GPT-3.5 Turbo}, attaining a Precision@1 of 0.8114, outperforming other baselines, including IReRa (0.7049), LLM4Jobs (0.6644 for CoT), and LC (0.634).

A one-way ANOVA indicated no significant differences between the LLMs ($p > 0.05$). These results demonstrate the robustness of the framework, which consistently produces competitive performance across different LLMs despite their taxonomic knowledge gaps (as shown in Table~\ref{tab:assessing_performance_all}). Even less knowledgeable models like Llama 3 8B Instruct and Mistral Small, achieved comparable Precision@1 scores using the framework.

Overall, methods using TGRE significantly outperform CoT ($p < 0.05$), demonstrating the effectiveness of the grounded rationale approach. The post-hoc test shows that TGRE with sentence-based retrieval significantly outperforms CoT with label-based retrieval ($p < 0.05$), with no significant differences between other method and retrieval unit combinations. Furthermore, the sentence-based retrieval unit outperforms the label-based unit in precision ($p < 0.05$), highlighting its effectiveness in leveraging richer contextual information.

Next, there are significant differences in the performance of TGRE compared to its ablated variants, TGRE- and TGRE-\,- ($p < 0.001$), and between CoT and its ablated variants, COT- and COT-\,- ($p < 0.001$). These further underscore the importance of retrieval and reranking. The absence of reranking leads to a notable drop in performance.
Removing both retrieval and reranking results in the steepest decline.
Results for the inference-only variants (TGRE-\,- and CoT-\,-) are consistent with the taxonomic knowledge of each LLM, where more knowledgeable models greatly outperform less knowledgeable models. 

In summary, the findings demonstrate the effectiveness of the proposed framework, which combines LLM reasoning capabilities with taxonomy-guided examples, retrieval, and reranking to achieve robust performance across efficient LLMs in occupation classification tasks. Beyond performance improvements, the computational cost analysis in Appendix~\ref{sec:cost_analysis} indicates that the framework is highly cost-effective. While the framework incurs marginal processing time due to retrieval and reranking, our analysis shows that it substantially reduces API costs by 72\% compared to direct GPT-4o inference, making it a more practical and scalable approach for large-scale classification tasks.

\section{Skill Classification Evaluation}
\label{sec:skill_evaluation}

To demonstrate the versatility of the proposed framework, we extended it to a multi-label skill classification task. This task involves assigning one or more skills and competences from the ESCO taxonomy (ESCO) to a given job vacancy post. Similar to O*NET-SOC, ESCO is the European standard classification system for Skills, Competences and Occupations. In this study, we employ the \textbf{ESCO v1.2} taxonomy, which contains 13,939 skills and competences (class labels).

By adapting the framework to this task, the input space $\mathbb{X}$ consists of job vacancy posts, the label spaces $\mathbb{Y}$ comprises skills and competences from the ESCO taxonomy $\mathbb{T}$. During the \textbf{inference} stage, the framework processes each job vacancy post ($x$) to infer one or more skills or competences ($y' \subseteq \mathbb{Y}$) relevant to the post. For TGRE, in-context examples are augmented with skill descriptions from ESCO. For CoT, the LLM generates chain-of-thought rationales for the in-context examples. In the \textbf{retrieval} stage, the inferred set of labels ($y'$) and/or rationales ($r'$) are used as queries to retrieve a set of candidate skills and competences from ESCO. Finally, the \textbf{reranking} stage refines the retrieved candidates using the LLM to evaluate their relevant to $x$. 

\subsection{Dataset}
\label{subsec:skill_dataset}

We used a labeled dataset \cite{decorte2022design,zhang2022skillspan} of technology-sector job vacancy posts, where skill candidates for each post were manually annotated with corresponding ESCO skills and competences. To ensure reproducibility, we adopted the same test and validation splits as those used by D'Oosterlinck et al. \cite{d2024context}. Duplicate skills and competences were removed from each job post in both sets to improve data quality. The test set includes 338 annotated job posts and 272 unique ground-truth labels, while the validation set, comprising 75 annotated posts and 67 unique ground-truth labels, was employed to supply in-context examples. On average, job posts in the test and validation sets contain 12.9 words (S.D. = 8.7) and 12 words (S.D. = 10.1), respectively. Each job post in the test set is associated with an average of 1.9 ground-truth labels (S.D. = 1.4), while those in the validation set have an average of 2 labels (S.D. = 1.3). 

\subsection{Experimental Setup}
\label{subsec:skill_setup}

We adopted the same experimental setup as described in Section~\ref{subsec:occupation_setup}, including LLM models and temperature setting, reasoning methods, retrieval units, and ablation variants. The LLMs were also accessed during the same periods as described in the occupation classification evaluation. All prompt templates and in-context examples are provided in Section~\ref{subsec:skill_prompt_templates} of the appendix.

To handle the extensive label space in the ESCO taxonomy, we set the number of top candidates retrieved during the retrieval stage to $m = 50$, following D'Oosterlinck et al. \cite{d2024context}. In the reranking stage, the number of top-$n$ candidates was set to $n = 10$. The number of in-context examples for both TGRE and CoT reasoning methods during inference and reranking was set to $k = 2$. For CoT, these in-context examples were selected by the optimization module in IReRa \cite{d2024context}, whereas they were randomly chosen in TGRE-based variants.

To enable direct comparison with D'Oosterlinck et al. \cite{d2024context}, we introduced additional GPT-4-based variants (\textbf{TGRE*} and \textbf{CoT*}) for each top TGRE and CoT performer, using GPT-4o as the LLM in the reranking stage. These variants also serve as an upper bound for performance comparison. In addition, we report the performance of the best fine-tuned systems on the same dataset by Decorter et al. \cite{decorte2023extreme} and Clavi{\'e} et al. \cite{clavie2023large}, which were taken directly from their papers.

\begin{table*}[!ht]
\caption{RP@10 for skill classification. The best results are highlighted in bold. The columns abbreviate the full names of the LLMs: GPT refers to GPT-3.5 Turbo, Gemini to Gemini-1.5 Flash, Claude to Claude Instant 1.2, Llama to Llama 3 8B Instruct, and Mistral to Mistral Small. For TGRE* and CoT*, we use GPT-4o for reranking.}
\label{tab:skills_results_10}
\centering
\begin{tabular}{@{}lllrrrrr@{}}
\toprule
Baseline & Reasoning & Retrieval & \multicolumn{1}{l}{GPT} & \multicolumn{1}{l}{Gemini} & \multicolumn{1}{l}{Claude} & \multicolumn{1}{l}{Llama} & \multicolumn{1}{l}{Mistral} \\ 
Equivalence & Method & Unit &  & & & & \\ \midrule
& TGRE* & Sentence & \textbf{0.7045} & \textbf{0.7052} & \textbf{0.6959} & \textbf{0.7023} & \textbf{0.6929} \\
& TGRE & Sentence & 0.6331 & 0.6557 & 0.6202 & 0.5937 & 0.62 \\
& CoT & Sentence & 0.2466 & 0.2909 & 0.3968 & 0.2862 & 0.3105 \\
\midrule
& TGRE & Label & 0.5744 & 0.6251 & 0.5811 & 0.4825 & 0.5624 \\
IReRa & CoT* & Label & 0.645 & 0.6546 & 0.6338 & 0.5704 & 0.6149 \\
IReRa & CoT & Label & 0.5993 & 0.5881 & 0.5381 & 0.4216 & 0.5113 \\
\midrule
LLM4Jobs & TGRE- & Sentence & 0.5968 & 0.6031 & 0.6086 & 0.6094 & 0.5904 \\
LLM4Jobs & CoT- & Sentence & 0.181 & 0.1457 & 0.2637 & 0.1767 & 0.2067 \\
 & TGRE- & Label & 0.5638 & 0.6151 & 0.5703 & 0.4705 & 0.5404 \\
 & CoT- & Label & 0.5717 & 0.5224 & 0.4452 & 0.354 & 0.4506 \\
 & TGRE-\,- & None & 0.1527 & 0.1045 & 0.1164 & 0.0993 & 0.1008 \\
 & CoT-\,- & None & 0.0325 & 0.0592 & 0.028 & 0.0163 & 0.0466 \\ \bottomrule
\end{tabular}
\end{table*}

\subsection{Evaluation Metric}
\label{subsec:skill_metric}

We employed \textbf{R-precision@}\textit{K} (\textbf{RP@}\textit{K}), where \textit{K} = 3, 5, and 10, to evaluate multi-label classification performance. Unlike Precision@\textit{K}, RP@\textit{K} measures the proportion of correctly classified labels among the top-\textit{K} predictions, normalized by the minimum of either the total number of predicted labels or the total number of ground-truth labels for a given job post. This metric, commonly used in skill classification tasks \cite{decorte2023extreme,clavie2023large,d2024context}, accounts for label variability per instance. We performed the same statistical analyses as in occupation classification to assess significant differences in results.

\subsection{Results}
\label{subsec:skill_results}

Table~\ref{tab:skills_results_10} presents the results for RP@10 across methods and models. Due to space limitations, RP@3 and RP@5 results are provided in Tables~\ref{tab:skills_results_3} and ~\ref{tab:skills_results_5} in Section ~\ref{sec:supp_tables} of the appendix. The best performance is achieved by \textbf{TGRE* with sentence-based retrieval}, attaining an RP@10 of 0.7052 on \textbf{Gemini-1.5 Flash}, outperforming IReRa (0.6546)\footnote{Note that the state-of-the-art RP@10 of 0.7058 was achieved by the best IReRa variant as reported in their original experiment \cite{d2024context} though we were unable to independently replicate this result.} and LLM4Jobs (0.6151). CoT* also greatly outperforms the best CoT variant across all models. Consistent with previous findings \cite{clavie2023large,d2024context}, these results reinforce the effectiveness of GPT-4 as a powerful reranker for skill classification. 

Compared to fine-tuned systems, the best TGRE* with sentence-based retrieval achieved an RP@5 score of 0.6318, exceeding the RP@5 scores reported by Decorte et al. \cite{decorte2023extreme} (0.5462) and Clavi{\'e} et al. \cite{clavie2023large} (0.615). Meanwhile, the best TGRE with sentence-based retrieval achieved an RP@5 score of 0.5681, outperforming Decorte et al.'s system, as shown in Tables~\ref{tab:finetuning_score} and ~\ref{tab:skills_results_5} in the appendix.

In line with the findings from the occupation classification task, no significant differences were found between the performance of the various LLMs ($p > 0.05$), indicating that all models are equally effective for this task. Next, TGRE significantly outperforms CoT across various LLMs and retrieval units ($p < 0.001$), underscoring the benefit of incorporating taxonomy-guided reasoning. The post-hoc test indicates that most method and LLMs combinations do not significantly differ, with a few exceptions: TGRE on Claude Instant significantly outperforms CoT on Llama 3 ($p < 0.05$), TGRE on Gemini-1.5 significantly outperforms CoT on Llama 3 ($p < 0.01$), and TGRE on Gemini-1.5 significantly outperforms CoT on Mistral ($p < 0.05$). 

Additionally, the post-hoc analysis shows significant differences between most method and retrieval unit combinations ($p < 0.01$), except for the comparison between TGRE and CoT when using label-based retrieval ($p > 0.05$). These results highlight the robustness of TGRE across different domains, as demonstrated by its consistent superiority in both occupation and skill classification tasks.

Lastly, sentence-based retrieval units significantly outperform label-based units across most configurations ($p < 0.05$), highlighting the importance of leveraging rich taxonomic context. For instance, TGRE with sentence-based retrieval on GPT-3.5 Turbo achieves an RP@10 of 0.7045, while label-based retrieval achieves only 0.5744. Consistent with the occupation classification results,  significant differences in performance were found between TGRE and its ablated variants ($p < 0.001$), and between CoT and its ablated variants ($p < 0.001$). These findings underscore the importance of both retrieval and reranking  across domains.

Overall, these results validate the effectiveness and versatility of the proposed framework. By combining taxonomy-guided reasoning, retrieval, and reranking, the framework achieves strong and robust performance in both multi-class occupation classification and multi-label skill classification, demonstrating its adaptability across domains. Although we did not directly assess LLM knowledge of ESCO, the lower performance scores and task complexity in the skill classification may reflect a more limited knowledge of ESCO compared to O*NET-SOC, underscoring the importance of the framework in bridging the internal knowledge gaps.

\section{Limitations and Future Work}
\label{sec:limitations}

\textbf{Knowledge assessment}: This study focused exclusively on the ONET-SOC taxonomy to assess LLM knowledge, a choice justified by its widespread use. Expanding future studies to include other taxonomies, such as ESCO or ISCO, could improve generalizability.

\textbf{Framework}: While simplicity and adaptability are key strengths of the framework, there remain areas of improvement. First, some manual prompt engineering is still required for adapting the framework to other domains. Next, different formatting function $f$ could be designed for incorporating taxonomic knowledge into grounded rationales. Additionally, future work could explore ways to handle flat taxonomies or dictionaries, which lack hierarchical structures or rich class information, e.g., generating auxiliary data with LLMs to enrich such taxonomies or leveraging examples from similar taxonomies. Furthermore, we employed fixed \textit{k}-shot in-context examples without dynamically selecting the optimal \textit{k} examples or utilizing advanced selection strategies, such as k-nearest neighbor based methods \cite{liu2021makes}. These strategies present promising directions for improvement. Similarly, future work could explore different choices for top-\textit{m} retrieval and top-\textit{n} reranking more thoroughly.

\textbf{Evaluation}: GPT-4 based evaluator used in occupation classification is not entirely free from bias. The manual validation of GPT-4o relied on a small human-annotated sample and scaling up these annotations could solidify findings. Similarly, the annotated SkillSpan dataset used for skill classification is relatively small. Using larger and more diverse datasets would improve the robustness of evaluation. For occupation classification, non-LLM or fine-tuned baselines were not included due to the lack of labeled training data. Future work could explore synthetic data generation to enable the fine-tuning of robust classification models.

\section{Conclusion and Implications}
\label{sec:conclusion}

In this paper, we investigate the capabilities of large language models (LLMs) in understanding the ONET-SOC taxonomy. The study begins with a knowledge assessment experiment to evaluate the performance of both frontier and efficient LLMs in generating precise occupational titles and codes. The results indicate that frontier models, especially GPT-4o, consistently outperform the efficient ones, demonstrating their in-depth taxonomic understanding. However, LLM performance varies across different levels of code granularity within the taxonomy, as evidenced by their lower performance in generating specific occupations compared to major occupations. 

To address these challenges, we propose a multi-stage framework for occupation classification that incorporates inference, retrieval, and reranking stages. Central to our approach is the taxonomy-guided reasoning examples that integrates grounded rationales from the taxonomy to improve classification accuracy. We apply the framework to both occupation and skill classification tasks, demonstrating its versatility and effectiveness. Our experiments show that the framework produces competitive performance across different LLMs, regardless of their underlying taxonomic knowledge.

This study offers several significant implications. Firstly, our analysis highlights the limitations of chain-of-thought (CoT) reasoning, particularly in tasks requiring domain-specific knowledge. While CoT is effective at general problem solving, our results show that it under-performs when deeper domain-specific understanding is necessary. Secondly, our findings highlight the limitations of LLM knowledge for occupation classification when directly inferring SOC labels. This underscores the crucial role of leveraging external taxonomic information to enhance the classification accuracy. By incorporating detailed class descriptions from taxonomies, LLMs generated better aligned queries, which in turn improved retrieval and reranking. This demonstrates how taxonomic knowledge can effectively address the gaps in LLMs' understanding, particularly in specialized classification tasks. The effectiveness of this approach highlights the potential for extending the use of taxonomic or domain-specific knowledge to related classification problems. Lastly, the study contributes to computational social science (CSS) by addressing key challenges in automatic content analysis. The proposed framework offers an accessible and adaptable solution for CSS domains, which often face data scarcity, difficulties in data collection, and a reliance on domain-specific knowledge and taxonomies.

\section{Acknowledgments}
\label{sec:acknowledgements}

This research / project is supported by the Ministry of Education, Singapore, under its MOE Academic Research Fund Tier 2 programme (Award T2EP20223-0047). Any opinions, findings and conclusions or recommendations expressed in this material are those of the author(s) and do not reflect the views of the Ministry of Education, Singapore. The authors would also like to acknowledge support from the Data Science Institute (DSI) at Columbia University and thank the anonymous reviewers for their constructive feedback, which significantly improved the quality of this work.

\section{Code and Data Availability}
The code and datasets are publicly available at \url{https://github.com/aekpalakorn/TGRE-Classification}.

\bibliography{main}

\begin{thebibliography}{24}
\providecommand{\natexlab}[1]{#1}

\bibitem[{Boselli et~al.(2017)Boselli, Cesarini, Mercorio, and Mezzanzanica}]{boselli2017using}
Boselli, R.; Cesarini, M.; Mercorio, F.; and Mezzanzanica, M. 2017.
\newblock Using machine learning for labour market intelligence.
\newblock In \emph{ECML PKDD 2017}, 330--342. Springer.

\bibitem[{Clavi{\'e} and Souli{\'e}(2023)}]{clavie2023large}
Clavi{\'e}, B.; and Souli{\'e}, G. 2023.
\newblock Large language models as batteries-included zero-shot ESCO skills matchers.
\newblock \emph{arXiv preprint arXiv:2307.03539}.

\bibitem[{Decorte et~al.(2022)Decorte, Van~Hautte, Deleu, Develder, and Demeester}]{decorte2022design}
Decorte, J.-J.; Van~Hautte, J.; Deleu, J.; Develder, C.; and Demeester, T. 2022.
\newblock Design of negative sampling strategies for distantly supervised skill extraction.
\newblock \emph{arXiv preprint arXiv:2209.05987}.

\bibitem[{Decorte et~al.(2023)Decorte, Verlinden, Van~Hautte, Deleu, Develder, and Demeester}]{decorte2023extreme}
Decorte, J.-J.; Verlinden, S.; Van~Hautte, J.; Deleu, J.; Develder, C.; and Demeester, T. 2023.
\newblock Extreme multi-label skill extraction training using large language models.
\newblock \emph{arXiv preprint arXiv:2307.10778}.

\bibitem[{D'Oosterlinck et~al.(2024)D'Oosterlinck, Khattab, Remy, Demeester, Develder, and Potts}]{d2024context}
D'Oosterlinck, K.; Khattab, O.; Remy, F.; Demeester, T.; Develder, C.; and Potts, C. 2024.
\newblock In-context learning for extreme multi-label classification.
\newblock \emph{arXiv preprint arXiv:2401.12178}.

\bibitem[{Guo et~al.(2024)Guo, Handzlik, Jones, and Skiena}]{guo2024evolution}
Guo, X.; Handzlik, D.; Jones, J.~J.; and Skiena, S.~S. 2024.
\newblock The evolution of occupational identity in twitter biographies.
\newblock In \emph{Proceedings of ICWSM 2024}, volume~18, 502--514.

\bibitem[{Hu et~al.(2016)Hu, Xiao, Luo, and Nguyen}]{hu2016language}
Hu, T.; Xiao, H.; Luo, J.; and Nguyen, T.-v.~T. 2016.
\newblock What the language you tweet says about your occupation.
\newblock In \emph{Proceedings of ICWSM 2016}, volume~10, 181--190.

\bibitem[{Javed et~al.(2015)Javed, Luo, McNair, Jacob, Zhao, and Kang}]{javed2015carotene}
Javed, F.; Luo, Q.; McNair, M.; Jacob, F.; Zhao, M.; and Kang, T.~S. 2015.
\newblock Carotene: A Job Title Classification System for the Online Recruitment Domain.
\newblock In \emph{Proceedings of IEEE BigDataService 2015}, 286--293.

\bibitem[{Kalhor et~al.(2024)Kalhor, Gardner, Weber, and Kashyap}]{kalhor2024gender}
Kalhor, G.; Gardner, H.; Weber, I.; and Kashyap, R. 2024.
\newblock Gender Gaps in Online Social Connectivity, Promotion and Relocation Reports on LinkedIn.
\newblock In \emph{Proceedings of ICWSM 2024}, volume~18, 800--812.

\bibitem[{Khattab et~al.(2023)Khattab, Singhvi, Maheshwari, Zhang, Santhanam, Vardhamanan, Haq, Sharma, Joshi, Moazam et~al.}]{khattab2023dspy}
Khattab, O.; Singhvi, A.; Maheshwari, P.; Zhang, Z.; Santhanam, K.; Vardhamanan, S.; Haq, S.; Sharma, A.; Joshi, T.~T.; Moazam, H.; et~al. 2023.
\newblock Dspy: Compiling declarative language model calls into self-improving pipelines.
\newblock \emph{arXiv preprint arXiv:2310.03714}.

\bibitem[{Kumar et~al.(2024)Kumar, Kim, Ravi, Sun, Faloutsos, Salakhutdinov, and Yoon}]{kumar2024automatic}
Kumar, R.; Kim, Y.; Ravi, S.; Sun, H.; Faloutsos, C.; Salakhutdinov, R.; and Yoon, M. 2024.
\newblock Automatic question-answer generation for long-tail knowledge.
\newblock \emph{arXiv preprint arXiv:2403.01382}.

\bibitem[{Li, Kang, and De~Bie(2023)}]{li2023llm4jobs}
Li, N.; Kang, B.; and De~Bie, T. 2023.
\newblock LLM4Jobs: Unsupervised occupation extraction and standardization leveraging Large Language Models.
\newblock \emph{arXiv preprint arXiv:2309.09708}.

\bibitem[{Liu et~al.(2021)Liu, Shen, Zhang, Dolan, Carin, and Chen}]{liu2021makes}
Liu, J.; Shen, D.; Zhang, Y.; Dolan, B.; Carin, L.; and Chen, W. 2021.
\newblock What Makes Good In-Context Examples for GPT-$3 $?
\newblock \emph{arXiv preprint arXiv:2101.06804}.

\bibitem[{Liu et~al.(2023)Liu, Iter, Xu, Wang, Xu, and Zhu}]{liu2023g-eval}
Liu, Y.; Iter, D.; Xu, Y.; Wang, S.; Xu, R.; and Zhu, C. 2023.
\newblock {G}-Eval: {NLG} Evaluation using Gpt-4 with Better Human Alignment.
\newblock In \emph{Proceedings of EMNLP 2023}, 2511--2522.

\bibitem[{Luo et~al.(2023)Luo, Vu, Phung, and Haf}]{luo2023systematic}
Luo, L.; Vu, T.; Phung, D.; and Haf, R. 2023.
\newblock Systematic Assessment of Factual Knowledge in Large Language Models.
\newblock In \emph{Findings of the Association for Computational Linguistics: EMNLP 2023}, 13272--13286.

\bibitem[{Russ et~al.(2016)Russ, Ho, Colt, Armenti, Baris, Chow, Davis, Johnson, Purdue, Karagas et~al.}]{russ2016computer}
Russ, D.~E.; Ho, K.-Y.; Colt, J.~S.; Armenti, K.~R.; Baris, D.; Chow, W.-H.; Davis, F.; Johnson, A.; Purdue, M.~P.; Karagas, M.~R.; et~al. 2016.
\newblock Computer-based coding of free-text job descriptions to efficiently identify occupations in epidemiological studies.
\newblock \emph{Occupational and environmental medicine}, 73(6): 417--424.

\bibitem[{Soroush et~al.(2024)Soroush, Glicksberg, Zimlichman, Barash, Freeman, Charney, Nadkarni, and Klang}]{soroush2024large}
Soroush, A.; Glicksberg, B.~S.; Zimlichman, E.; Barash, Y.; Freeman, R.; Charney, A.~W.; Nadkarni, G.~N.; and Klang, E. 2024.
\newblock Large language models are poor medical coders—benchmarking of medical code querying.
\newblock \emph{NEJM AI}, 1(5): AIdbp2300040.

\bibitem[{Sottana et~al.(2023)Sottana, Liang, Zou, and Yuan}]{sottana2023evaluation}
Sottana, A.; Liang, B.; Zou, K.; and Yuan, Z. 2023.
\newblock Evaluation Metrics in the Era of GPT-4: Reliably Evaluating Large Language Models on Sequence to Sequence Tasks.
\newblock In \emph{Proceedings of EMNLP 2023}, 8776--8788.

\bibitem[{Sun et~al.(2023)Sun, Xu, Zha, Liu, and Dong}]{sun2023head}
Sun, K.; Xu, Y.~E.; Zha, H.; Liu, Y.; and Dong, X.~L. 2023.
\newblock Head-to-tail: How knowledgeable are large language models (llm)? AKA will llms replace knowledge graphs?
\newblock \emph{arXiv preprint arXiv:2308.10168}.

\bibitem[{Sun et~al.(2024)Sun, Xin, Sun, Xu, Yang, Dong, Tang, and Chen}]{sun2024large}
Sun, Y.; Xin, H.; Sun, K.; Xu, Y.~E.; Yang, X.; Dong, X.~L.; Tang, N.; and Chen, L. 2024.
\newblock Are Large Language Models a Good Replacement of Taxonomies?
\newblock \emph{Proc. VLDB Endow.}, 17(11): 2919–2932.

\bibitem[{T{\"a}nzer, Ruder, and Rei(2022)}]{tanzer2022memorisation}
T{\"a}nzer, M.; Ruder, S.; and Rei, M. 2022.
\newblock Memorisation versus Generalisation in Pre-trained Language Models.
\newblock In \emph{Proceedings of ACL 2022}, 7564--7578.

\bibitem[{Wei et~al.(2022)Wei, Wang, Schuurmans, Bosma, Xia, Chi, Le, Zhou et~al.}]{wei2022chain}
Wei, J.; Wang, X.; Schuurmans, D.; Bosma, M.; Xia, F.; Chi, E.; Le, Q.~V.; Zhou, D.; et~al. 2022.
\newblock Chain-of-thought prompting elicits reasoning in large language models.
\newblock \emph{NeurIPS}, 35: 24824--24837.

\bibitem[{Zhang et~al.(2022)Zhang, Jensen, Sonniks, and Plank}]{zhang2022skillspan}
Zhang, M.; Jensen, K.; Sonniks, S.; and Plank, B. 2022.
\newblock SkillSpan: Hard and Soft Skill Extraction from English Job Postings.
\newblock In \emph{Proceedings of NAACL 2022}, 4962--4984.

\bibitem[{Zhang, Jensen, and Plank(2022)}]{zhang2022kompetencer}
Zhang, M.; Jensen, K.~N.; and Plank, B. 2022.
\newblock Kompetencer: Fine-grained skill classification in danish job postings via distant supervision and transfer learning.
\newblock In \emph{Proceedings of LREC 2022}, 436--447.

\end{thebibliography}


\section{Ethical Considerations}
\label{sec:ethical}

In this section, we discuss several ethical considerations relevant to our research to ensure the responsible use and development of automatic occupation classification systems.

\textbf{Model and taxonomic biases}: The LLMs utilized in this study inherently exhibit biases from their pretraining data, which can lead to inaccurate or unfair classifications. In addition, occupational taxonomies, such as O*NET-SOC, may reflect inherent biases in how occupations are defined and categorized, particularly since the O*NET-SOC taxonomy is US-centric. These biases can reinforce stereotypes or perpetuate inequalities, especially when certain occupations are associated with race, gender, or socioeconomic factors. Future work should focus on systematically addressing these biases to ensure equitable outcomes.

\textbf{Unintended consequences of automated systems}: Automated classification systems, including those developed in this study, may result in unintended consequences if misused or applied without human supervision. Even minor errors can lead to significant downstream effects. For example, a hiring platform relying on automated occupation classification systems might overestimate their accuracy, A misclassification could lead to a candidate's resume being assigned an incorrect occupation, potentially impacting their employment opportunities. To mitigate such risks, these systems should be used to complement rather than fully replace human judgment.

\textbf{Access equity}: Our framework is designed to be lightweight and model-agnostic, allowing it to be implemented across a range of LLMs, including efficient models. This reduces barriers to adoption for researchers and organizations with limited computational resources. However, a gap remains between high-resource and low-resource organizations, particularly in accessing high-performance models.  Bridging this gap will require broader access to high-performing LLMs and open-source alternatives.

\textbf{Environmental impact}: Although our framework is training-free and focuses on leveraging existing LLMs, model inference still requires GPU usage and contributes to the carbon footprint. By emphasizing the use of efficient LLMs, our research mitigates but does not fully eliminate this impact. To further minimize environmental impact, further efforts in optimizing LLM efficiency are needed. 

\appendix

\section{The Lightcast Dataset}
\label{sec:lightcast}
The Lightcast dataset, obtained in October 2022, comprises approximately 141.6 million anonymized US-based resume profiles collected from various online professional networking platforms by Lightcast, a labor market analytics company. Each profile is enriched with occupation (SOC title and code), industry, and education details through machine learning-based standardization and processing\footnote{https://kb.lightcast.io/en/articles/7153977-global-data-101}, and comprises job history, education history, and skills sections. To manage the dataset size, we partition profiles using a hashing-based method on their unique identifiers, resulting in 284 partitions with approximately 458,000 profiles each. These partitions, labeled ``LC'' followed by their index (e.g., LC0 to LC283), ensure a random and uniform distribution of the profile data.

To ensure data quality and authenticity, we performed data cleaning by applying these criteria: (i) each job record in the profile must include the job title, company name, city, state, country, and start date. Past job records also required an end date. Job records with end dates preceding start dates were excluded; (ii) job titles in the profile containing non-occupational terms (``student,'' ``intern,'' or ``owner'') as well as those indicating multiple roles within one title were excluded; and (iii) each education record in the profile must specify the degree obtained, start date, end date, and school name.

To construct the Jobs12K test set, we applied the following steps to sample 11,920 job records from the job histories of all profiles in the LC0 partition: (i) exclude job titles with non-alphanumeric, non-ASCII characters, or more than seven words; and (iii) For the 596 most common 6-digit SOC codes (those linked to at least 20 records), randomly sample 20 job records per code, resulting in 11,920 records.

Popular occupations used in the selection may not represent the true occupational distribution, as default SOC codes provided by Lightcast may lack precision at the 6-digit level. Instead, these codes indicate a preliminary signal of commonly represented occupations.

\section{The Formatting Functions}
\label{sec:formatting_functions}

In our proposed TGRE framework, the formatting function \textit{f} is a component that converts taxonomic information retrieved for a specific class label \textit{y} into a grounded rationale \textit{r} that aligns the LLM reasoning process with the taxonomy. The rationale \textit{r} serves as an intermediate explanation that clarifies the connection between the input \textit{x} and the corresponding label \textit{y}. For example, it explains how a job title-company pair is related to an SOC title and code.

Formally, \textit{f} takes the retrieved taxonomic information $T(y)$, which consists of information of \textit{y}, (e.g., occupation description), and transforms it into a rationale $r=f(T(y))$. The rationale is then embedded into in-context examples to guide the reasoning process of LLMs.

In this study, we use a simple implementation of \textit{f} where the retrieved taxonomic information $T(y)$ is incorporated verbatim in the rationale. This approach minimizes the framework's complexity while achieving strong empirical results. For example:

Given an input \textit{x} ``records and evidence technician, cobb county police department'' and its SOC label \textit{y} ``Police Identification and Records Officers (33-3021.02)'' in the validation set, the proposed framework retrieves the matched taxonomic entity $T(y)$ from the O*NET-SOC taxonomy, comprising the followings:

\begin{itemize}
    \item Occupation title: Police Identification and Records Officers
    \item Occupation code: 33-3021.02
    \item Description: Collect evidence at crime scenes, classify and identify fingerprints, and photograph evidence for use in criminal and civil cases.
\end{itemize}

The verbatim formatting function retrieves the description from $T(y)$ and directly into the grounded rationale \textit{r} as follows:

\begin{itemize}
    \item Grounded rationale: Given the job title '\textbf{records and evidence technician}' and company '\textbf{cobb county police department},' the individual is likely expected to \textbf{collect evidence at crime scenes, classify and identify fingerprints, and photograph evidence for use in criminal and civil cases}.
\end{itemize}

Here, the elements in bold are values derived from the task inputs and taxonomy that replace placeholders within the rationale template.

It is important to note that while we employ a simple verbatim approach in this study, LLMs can be leveraged to make more sophisticated formatting functions. For example, an LLM could rewrite the retrieved description for improved brevity and clarity. Moreover, it could incorporate other taxonomic information beyond descriptions, such as tasks and skills, to generate an enriched summary.

\section{Optimization Module in IReRa}
\label{sec:irera_optimization}

The optimization module in IReRa, as depicted in Figure 1 in \citet{d2024context}, employs bootstrapping to iteratively identify the most effective few-shot in-context examples from the validation set. This procedure involves systematically evaluating the performance of various combinations of examples and selecting those that achieve the best results based on a predefined metric. In our experiments, this optimization process identified a specific 1-shot in-context example ($k=1$) for the CoT-based variant in the occupation classification experiment. For the skill classification experiment, the optimization module selected a set of 2-shot in-context examples ($k=2$) for the CoT-based variant. While our TGRE variant also used $k=2$, the examples were randomly chosen, distinguishing them from the optimized examples used by the CoT-based variant.

\section{Prompt Templates and In-context Examples}
\label{sec:prompt_templates}
This appendix provides detailed prompt templates and in-context examples used across different stages of the study

\subsection{Knowledge Assessment}
\label{subsec:knowledge_prompt_templates}
In this section, we provide the prompt templates designed to evaluate the LLMs' ability to generate SOC titles and codes based on the O*NET-SOC taxonomy. Figure~\ref{prompt:knowledge} displays recall and recognition prompt templates for the knowledge assessment tasks. Variables in the form \$\{$var\_name$\}, highlighted in blue, are replaced with corresponding values from O*NET-SOC.

\begin{figure}[!h]
\centering
\begin{tcolorbox}[colback=blue!5!white, colframe=blue!75!black, title=Recall Prompt Template]
\small
\begin{VerbatimWrap}
Fill in the <blank>.
---
Output format: 
Generate either a canonical SOC title or code depending on the context. 
Encapsulate the answer in <answer></answer>. 
No other verbosity. Generate 0 if there is no suitable answer.
---
Example:
In the O*NET-SOC 2019 taxonomy, the \textcolor{blue}{\$\{example_answer_field\}} for the \textcolor{blue}{\$\{example_query_field\}} \textcolor{blue}{\$\{example_query\}} is <blank>.
<answer>\textcolor{blue}{\$\{example_answer\}}</answer>
---
In the O*NET-SOC 2019 taxonomy, the \textcolor{blue}{\$\{answer_field\}} for the \textcolor{blue}{\$\{query_field\}} \textcolor{blue}{\$\{query\}} is <blank>.
\end{VerbatimWrap}
\end{tcolorbox}
\begin{tcolorbox}[colback=blue!5!white, colframe=blue!75!black, title=Recognition Prompt Template]
\small
\begin{VerbatimWrap}
Select the best option to match entities in the O*NET-SOC 2019 taxonomy.
---
Output format: 
Encapsulate the answer in <answer></answer>. 
No other verbosity. Generate 0 if there is no suitable answer. 
---
Example:
Which one of the options is the \textcolor{blue}{\$\{example_answer_field\}} of \textcolor{blue}{\$\{example_query_field\}} \textcolor{blue}{\$\{example_query\}}?
Options: 
\textcolor{blue}{\$\{ex_options\}}
<answer>\textcolor{blue}{\$\{example_answer\}}</answer>
---
Which one of the options is the \textcolor{blue}{\$\{answer_field\}} for the \textcolor{blue}{\$\{query_field\}} \textcolor{blue}{\$\{query\}}?
Options: 
\textcolor{blue}{\$\{options\}}
\end{VerbatimWrap}
\end{tcolorbox}
\caption{Prompt templates for knowledge recall and recognition tasks. Variables in the form \$\{$var\_name$\}, highlighted in blue, are replaced with corresponding values from the O*NET-SOC taxonomy.}
\label{prompt:knowledge}
\end{figure}

\subsection{Occupation Classification}
\label{subsec:occupation_prompt_templates}
This section describes the prompt templates and in-context examples used in the inference and reranking stages of our framework for the occupation classification task. Figures ~\ref{prompt:tgre_occupation} and ~\ref{prompt:cot_occupation} display TGRE-based and CoT-based prompt templates for occupation classification, respectively. Figures ~\ref{examples:tgre_occupation} and ~\ref{examples:cot_occupation} display \textit{k}-shot in-context examples for TGRE-based and CoT-based prompt templates for occupation classification, respectively. These in-context examples are integrated into the respective templates by replacing the placeholder \$\{examples\}.

It is important to note that the TGRE prompt templates in Figure~\ref{prompt:tgre_occupation} do not specify the version of the O*NET taxonomy, unlike the CoT prompt templates in Figure ~\ref{prompt:cot_occupation}. During the prompt engineering, we found that specifying the ONET version did not affect task performance, as LLMs generally lack the precision required to generate SOC titles and codes for specific O*NET-SOC versions. As a result, we opted to omit the version number from TGRE prompts, while keeping them in the CoT prompts as the latter had already been used in prior experiments.

The output format for the CoT-based inference prompt template in Figure~\ref{prompt:cot_occupation} is an adaptation from IReRa \cite{d2024context}, originally designed for multi-label classification. To tailor this format to occupation classification, the output is truncated to retain only the top-1 predicted label.

\begin{figure}[!ht]
\centering
\begin{tcolorbox}[colback=blue!5!white, colframe=blue!75!black, title=TGRE-based Inference Prompt Template]
\small
\begin{VerbatimWrap}
Given a job title and a company name, infer the job description and predict the most likely O*NET-SOC occupational title and code. Only one SOC title and code per prediction. If insufficient context is given, specify "None" in the prediction. No other verbosity.
---
Follow the following format.
Reasoning: Given the job title "title" and company "company", the individual is likely responsible for...
Prediction: O*NET-SOC occupations
---
Examples:
\textcolor{blue}{\$\{examples\}}
---
Your task:
Job title and company: \textcolor{blue}{\$\{input\}}
\end{VerbatimWrap}
\end{tcolorbox}
\begin{tcolorbox}[colback=blue!5!white, colframe=blue!75!black, title=TGRE-based Reranking Prompt Template]
\small
\begin{VerbatimWrap}
Given a job title and a company name, select up to 10 most applicable O*NET-SOC occupational titles and codes to the job title and company name. If there is no applicable occupation, use the code "None (00-0000.00)." No other verbosity.
---
Follow the following format.
Reasoning: Given the job title "title"" and company "company", the individual is likely responsible for...
Most likely occupations: A ranked list of occupations, one occupation per line separated by new line.
1. Occupation A
2. Occupation B
---
Examples:
\textcolor{blue}{\$\{examples\}}
---
Your task:
Job title and company: \textcolor{blue}{\$\{input\}}
Options:
\textcolor{blue}{\$\{options\}}
\end{VerbatimWrap}
\end{tcolorbox}
\caption{TGRE-based prompt templates for occupation classification. Variables in the form \$\{$var\_name$\}, highlighted in blue, are replaced with corresponding values from task inputs}
\label{prompt:tgre_occupation}
\end{figure}

\begin{figure}[!ht]
\centering
\begin{tcolorbox}[colback=blue!5!white, colframe=blue!75!black, title=CoT-based Inference Prompt Template]
\small
\begin{VerbatimWrap}
Given a snippet of job title and company, predict O*NET-SOC 2019 occupation titles and codes it most likely belongs to. Only one SOC title and code per prediction. If insufficient context is given, specify "None" in the prediction. No other verbosity.
---
Follow the following format.
Reasoning: Let's think step by step in order to predict the O*NET 2019 occupation titles and codes it most likely belongs to. We ...
O*NET-SOC 2019 Occupations: list of SOC titles and 8-digit codes separated by semi-colons, e.g., Title 1 (11-1101.00); Title 2 (22-2220.22)
---
Examples:
\textcolor{blue}{\$\{examples\}}
---
Your task:
Job title and company: \textcolor{blue}{\$\{input\}}
\end{VerbatimWrap}
\end{tcolorbox}
\begin{tcolorbox}[colback=blue!5!white, colframe=blue!75!black, title=CoT-based Reranking Prompt Template]
\small
\begin{VerbatimWrap}
Given a snippet of job title and company, select up to 10 O*NET-SOC 2019 occupation titles and codes from the options most applicable to the job title and company name. Always include a numbered ranked list of O*NET-SOC 2019 Occupations. If there is no applicable occupation, use the code "None (00-0000.00)." No other verbosity.
---
Follow the following format.
Job title and company: Job title, company
Options: A list of O*NET-SOC occupational titles and codes to choose from
Reasoning: Let's think step by step in order to select up to 10 O*NET-SOC 2019 occupation titles and codes from the options most applicable to the job title and company name. We ...
O*NET-SOC 2019 Occupations: A ranked list of occupations, one occupation per line separated by new line.
1. Occupation A
2. Occupation B
...
---
Examples:
\textcolor{blue}{\$\{examples\}}
---
Your task:
Job title and company: \textcolor{blue}{\$\{input\}}
Options: 
\textcolor{blue}{\$\{options\}}
Reasoning: Let's think step by step in order to
\end{VerbatimWrap}
\end{tcolorbox}
\caption{CoT-based prompt templates for occupation classification. Variables in the form \$\{$var\_name$\}, highlighted in blue, are replaced with corresponding values from task inputs.}
\label{prompt:cot_occupation}
\end{figure}

\begin{figure*}[!ht]
\centering
\begin{tcolorbox}[colback=blue!5!white, colframe=blue!75!black, title=Examples for TGRE-based Inference Prompt]
\small
\begin{VerbatimWrap}
Job title and company: records and evidence technician, cobb county police department
Reasoning: Given the job title "records and evidence technician" and company "cobb county police department," the individual is likely expected to \textcolor{blue}{collect evidence at crime scene, classify and identify fingerprints, and photograph evidence for use in criminal and civil cases}.
Prediction: Police Identification and Records Officers (33-3021.02)
\end{VerbatimWrap}
\end{tcolorbox}
\begin{tcolorbox}[colback=blue!5!white, colframe=blue!75!black, title=Examples for TGRE-based Reranking Prompt]
\small
\begin{VerbatimWrap}
Job title and company: records and evidence technician, cobb county police department
Options:
Detectives and Criminal Investigators (33-3021.00); Forensic Science Technicians (19-4092.00); Police Identification and Records Officers (33-3021.02); Fraud Examiners, Investigators and Analysts (13-2099.04); Private Detectives and Investigators (33-9021.00); Intelligence Analysts (33-3021.06); Retail Loss Prevention Specialists(33-9099.02); Transportation Security Screeners (33-9093.00); Animal Control Workers (33-9011.00); Appraisers and Assessors of Real Estate (13-2023.00)
Reasoning: Given the job title "records and evidence technician" and company "cobb county police department," the individual is likely expected to \textcolor{blue}{collect evidence at crime scene, classify and identify fingerprints, and photograph evidence for use in criminal and civil cases}.
Most likely occupations:
1. Police Identification and Records Officers (33-3021.02)
2. Forensic Science Technicians (19-4092.00)
3. Detectives and Criminal Investigators (33-3021.00)
4. Fraud Examiners, Investigators and Analysts (13-2099.04)
5. Private Detectives and Investigators (33-9021.00)
6. Intelligence Analysts (33-3021.06)
7. Retail Loss Prevention Specialists (33-9099.02)
8. Transportation Security Screeners (33-9093.00)
9. Animal Control Workers (33-9011.00)
10. Appraisers and Assessors of Real Estate (13-2023.00)
\end{VerbatimWrap}
\end{tcolorbox}
\caption{In-context examples for TGRE-based prompts for occupation classification. Segments highlighted in blue are detailed description of the corresponding occupation retrieved from the taxonomy.}
\label{examples:tgre_occupation}
\end{figure*}

\begin{figure*}[!ht]
\centering
\begin{tcolorbox}[colback=blue!5!white, colframe=blue!75!black, title=Examples for CoT-based Inference Prompt]
\small
\begin{VerbatimWrap}
Job title and company: medical editor, medical editor and transcriptionist, zydoc medical transcription
Reasoning: Let's think step by step in order to predict the O*NET 2019 occupation titles and codes it most likely belongs to. We can start by breaking down the job title and company. "Medical editor" and "medical editor and transcriptionist" suggest that this person is responsible for editing and transcribing medical documents. The company, Zydoc Medical Transcription, confirms this assumption. Based on this information, we can predict that this job most likely belongs to the "Medical Transcriptionists" occupation (code 31-9094.00) in the O*NET 2019 database. This occupation involves transcribing medical reports and documents. Other possible occupations could include "Medical Records and Health Information Technicians" (code 29-2071.00) or "Editors"
O*NET-SOC 2019 Occupations: Medical Transcriptionists (31-9094.00); Medical Records and Health Information Technicians (29-2071.00); Editors (27-3041.00)
\end{VerbatimWrap}
\end{tcolorbox}
\begin{tcolorbox}[colback=blue!5!white, colframe=blue!75!black, title=Examples for CoT-based Reranking Prompt]
\small
\begin{VerbatimWrap}
Job title and company: medical editor, medical editor and transcriptionist, zydoc medical transcription
Options: 
Editors (27-3041.00); Medical Transcriptionists (31-9094.00); Medical Records Specialists (29-2072.00); Health Information Technologists and Medical Registrars (29-9021.00); Health Technologists and Technicians All Other (29-2099.00); Film and Video Editors (27-4032.00); Medical Equipment Repairers (49-9062.00); Emergency Medical Technicians (29-2042.00); Medical and Clinical Laboratory Technicians (29-2012.00); Health Informatics Specialists (15-1211.01)
Reasoning: Let's think step by step in order to select up to 10 O*NET-SOC 2019 occupation titles and codes from the options most applicable to the job title and company name. We can see that the job title includes "medical editor" and "medical transcriptionist", and the company name includes "medical transcription". This suggests that the job is likely related to medical editing and transcription. The options that best fit this description are Editors (27-3041.00) and Medical Transcriptionists (31-9094.00). Other options that could potentially fit are Health Information Technologists and Medical Registrars (29-9021.00) and Health Informatics Specialists (15-1211.01), as they also involve working with medical records and information. However, the job title does not mention any equipment repair or emergency medical services, so options such as Medical Equipment Repairers...
O*NET-SOC 2019 Occupations: 
1. Editors (27-3041.00) 
2. Medical Transcriptionists (31-9094.00) 
3. Health Information Technologists and Medical Registrars (29-9021.00) 
4. Health Informatics Specialists (15-1211.01)
\end{VerbatimWrap}
\end{tcolorbox}
\caption{In-context examples for CoT-based prompts generated by GPT-3.5 Turbo for occupation classification}
\label{examples:cot_occupation}
\end{figure*}

\begin{figure}[!ht]
\centering
\begin{tcolorbox}[colback=blue!5!white, colframe=blue!75!black, title=Zeroshot Occupation Selection Prompt]
\small
\begin{VerbatimWrap}
Given a job record (job title and company), select one or more O*NET-SOC labels (titles and codes) from the options which are most applicable to the job record. 
---
Follow this format:
Answer: List of applicable O*NET-SOC labels separated by semicolons. Each selected label should contain both title and code verbatim. If multiple labels are equally applicable, seperate them with semicolons in your answer. If there is no applicable label, select or indicate "None (00-0000.00)." No other verbosity.
---
Your task:
Job record: \textcolor{blue}{\$\{input\}}
Options:
\textcolor{blue}{\$\{options\}}
\end{VerbatimWrap}
\end{tcolorbox}
\caption{Zero-shot occupation selection prompt templates for occupation classification. Variables in the form \$\{$var\_name$\}, highlighted in blue, are replaced with corresponding values from task inputs.}
\label{prompt:zeroshot_annotate_occupation}
\end{figure}

\subsection{Skill Classification}
\label{subsec:skill_prompt_templates}

Here, we present the prompt templates and in-context examples used in the inference and reranking stages of our framework for the skill classification task. Figures ~\ref{prompt:tgre_skill} and ~\ref{prompt:cot_skill} display TGRE-based and CoT-based prompt templates for skill classification, respectively. Figures ~\ref{examples:tgre_infer_skill} - ~\ref{examples:cot_rank_skill_2} display \textit{k}-shot in-context examples for TGRE-based and CoT-based prompt templates for skill classification, respectively.

\begin{figure}[!ht]
\centering
\begin{tcolorbox}[colback=blue!5!white, colframe=blue!75!black, title=TGRE-based Inference Prompt Template]
\small
\begin{VerbatimWrap}
Given a snippet from a job vacancy, generate rationale in order to identify all ESCO skills explicitly mentioned. In the rationale, encapsulate ESCO concept and description in a square bracket.
---
Follow the following format.
Reasoning: The job vacancy snippet explicitly mentions...Therefore, we can infer the following ESCO skills and knowledge...[ESCO concept: description]...
---
Example #1:
\textcolor{blue}{\$\{example_1\}}
---
Example #2:
\textcolor{blue}{\$\{example_2\}}
---
Your task:
Text snippet: \textcolor{blue}{\$\{input\}}
\end{VerbatimWrap}
\end{tcolorbox}
\begin{tcolorbox}[colback=blue!5!white, colframe=blue!75!black, title=TGRE-based Reranking Prompt Template]
\small
\begin{VerbatimWrap}
Given a snippet from a job vacancy and a list of ESCO skills and knowledge as options, first generate a rationale then select verbatim up to 10 options most applicable to skills and knowledge explicitly mentioned in the snippet. No other verbosity.
---
Follow this out format.
Reasoning: The job vacancy snippet explicitly mentions...These phrases give us clues in determining the most relevant options by aligning them with the specific requirements in the snippet...
Therefore, we select the following 10 most applicable options: A ranked list of ESCO skills and knowledge, one skill per line separated by new line.
1. Skill A
2. Skill B 
...
---
Example #1:
\textcolor{blue}{\$\{example_1\}}
---
Example #2:
\textcolor{blue}{\$\{example_2\}}
---
Your task:
Text snippet: \textcolor{blue}{\$\{input\}}
Options:
\textcolor{blue}{\$\{options\}}
\end{VerbatimWrap}
\end{tcolorbox}
\caption{TGRE-based prompt templates for skill classification. Variables in the form \$\{$var\_name$\}, highlighted in blue, are replaced with corresponding values from task inputs.}
\label{prompt:tgre_skill}
\end{figure}

\begin{figure}[!ht]
\centering
\begin{tcolorbox}[colback=blue!5!white, colframe=blue!75!black, title=CoT-based Inference Prompt Template]
\small
\begin{VerbatimWrap}
Given a snippet from a job vacancy, identify all the ESCO job skills mentioned. Always return skills.
---
Follow the following format.
Vacancy: Job vacancy snippet
Reasoning: Let's think step by step in order to find the skills. We ...
Skills: list of comma-separated ESCO skills
---
\textcolor{blue}{\$\{example_1\}}
---
\textcolor{blue}{\$\{example_2\}}
---
Your task:
Vacancy: \textcolor{blue}{\$\{input\}}
Reasoning: Let's think step by step in order to
\end{VerbatimWrap}
\end{tcolorbox}
\begin{tcolorbox}[colback=blue!5!white, colframe=blue!75!black, title=CoT-based Reranking Prompt Template]
\small
\begin{VerbatimWrap}
Given a snippet from a job vacancy, generate a chain-of-thought rationale and select the 10 most applicable ESCO skills from the options that are directly expressed in the snippet.
---
Follow the following format.
Reasoning: Start the rationale with "Let's think step by step in order to identify the most applicable skills from the snippet. We..."
Skills: Ranked list of ESCO skills separated by semicolon
---
Example #1:
\textcolor{blue}{\$\{example_1\}}
---
Example #2:
\textcolor{blue}{\$\{example_2\}}
---
Your task:
Text snippet: \textcolor{blue}{\$\{input\}}
Options: 
\textcolor{blue}{\$\{options\}}
\end{VerbatimWrap}
\end{tcolorbox}
\caption{CoT-based prompt templates for skill classification. Variables in the form \$\{$var\_name$\}, highlighted in blue, are replaced with corresponding values from task inputs.}
\label{prompt:cot_skill}
\end{figure}

\begin{figure*}[!ht]
\centering
\begin{tcolorbox}[colback=blue!5!white, colframe=blue!75!black, title=Example 1 for TGRE-based Inference Prompt]
\small
\begin{VerbatimWrap}
Text snippet: You'll be required to apply your depth of knowledge and expertise to all aspects of the software development lifecycle as well as partner continuously with your many stakeholders on a daily basis to stay focused on common goals.
Reasoning: The job vacancy snippet explicitly mentions "software development lifecycle," "partner continuously with your many stakeholders," and "stay focused on common goals." Therefore, we can infer the following ESCO skills and knowledge:
- "Software development lifecycle" likely refers to the ESCO knowledge concept [systems development life-cycle: \textcolor{blue}{The sequence of steps, such as planning, creating, testing and deploying and the models for the development and life-cycle management of a system}].
- "Partner continuously with your many stakeholders" likely refers to the ESCO skill/competence concept [manage relationships with stakeholders: \textcolor{blue}{Create and maintain solid internal and external relations with stakeholders at operational level based on mutual trust and credibility in order to achieve organisational goals. Ensure organisational strategies incorporate strong stakeholder management and identify and prioritise strategic stakeholder relationships}].
- "Stay focused on common goals" likely refers to the ESCO skill/competence concept [identify with the company's goals: \textcolor{blue}{Act for the benefit of the company and for the achievement of its targets}].
\end{VerbatimWrap}
\end{tcolorbox}
\begin{tcolorbox}[colback=blue!5!white, colframe=blue!75!black, title=Example 2 for TGRE-based Inference Prompt]
\small
\begin{VerbatimWrap}
Text snippet: You are proficient in Python and English
Reasoning: The job vacancy snippet explicitly mentions "Python" and "English." Therefore, we can infer the following ESCO skills and knowledge:
- "Python" likely refers to the ESCO knowledge concept [Python (computer programming): \textcolor{blue}{The techniques and principles of software development, such as analysis, algorithms, coding, testing and compiling of programming paradigms in Python}].
- "English" likely refers to the ESCO knowledge concept [English: \textcolor{blue}{The English language}].
\end{VerbatimWrap}
\end{tcolorbox}
\caption{In-context examples for TGRE-based inference prompts for skill classification. Segments highlighted in blue are detailed description of the corresponding occupation retrieved from the taxonomy.}
\label{examples:tgre_infer_skill}
\end{figure*}

\begin{figure*}[!ht]
\centering
\begin{tcolorbox}[colback=blue!5!white, colframe=blue!75!black, title=Example 1 for TGRE-based Reranking Prompt]
\small
\begin{VerbatimWrap}
Text snippet: You'll be required to apply your depth of knowledge and expertise to all aspects of the software development lifecycle as well as partner continuously with your many stakeholders on a daily basis to stay focused on common goals.
Options:
Agile development; DevOps; Iterative development; Prototyping development; Rapid application development; Spiral development; Waterfall development; align efforts towards business development; collaborate on international energy projects; collaborate with coaching team; collaborate with designers; content development processes; cooperate to resolve information issues; cooperate with colleagues; cooperate with education professionals; create a work atmosphere of continuous improvement; create social alliances; develop production line; develop therapeutic relationships; engage with stakeholders; ensure cross-department cooperation; ensure cross-department cooperation; establish collaborative relations; focus on service; identify customer objectives; identify with the company's goals; incremental development; levels of software testing; liaise with cultural partners; liaise with engineers; maintain relationship with customers; maintain relationship with suppliers; manage personal progression; manage relationships with stakeholders; motivate employees; negotiate with stakeholders; negotiate with stakeholders; procurement lifecycle; product life-cycle; project commissioning; promote the principles of democracy and rule of law; research your target community; social alliances; solution deployment; strive for company growth; systems development life-cycle; talent management strategies; teamwork principles
Reasoning: The job vacancy snippet explicitly mentions "software development lifecycle," "partner continuously with your many stakeholders," and "stay focused on common goals." These phrases give us clues in determining the most relevant options by aligning them with the specific requirements in the snippet.
- The term "software development lifecycle" in the snippet directly corresponds to [systems development life-cycle: \textcolor{blue}{The sequence of steps, such as planning, creating, testing and deploying and the models for the development and life-cycle management of a system}], making it the most relevant option.
- The term "partner continuously with your many stakeholders" directly points to [manage relationships with stakeholders: \textcolor{blue}{Create and maintain solid internal and external relations with stakeholders at operational level based on mutual trust and credibility in order to achieve organisational goals. Ensure organisational strategies incorporate strong stakeholder management and identify and prioritise strategic stakeholder relationships}].
- The term "stay focused on common goals" suggests the importance of aligning individual efforts with the company's objectives, highlighting the need for employees to [identify with the company's goals: \textcolor{blue}{Act for the benefit of the company and for the achievement of its targets}].
Therefore, we select the following 10 most applicable options:
1. systems development life-cycle
2. manage relationships with stakeholders
3. identify with the company's goals
4. engage with stakeholders
5. negotiate with stakeholders
6. maintain relationship with customers
7. Waterfall development
8. Iterative development
9. Agile development
10. Rapid application development
\end{VerbatimWrap}
\end{tcolorbox}
\caption{In-context example 1 for TGRE-based reranking prompts for skill classification. Segments highlighted in blue are detailed description of the corresponding occupation retrieved from the taxonomy.}
\label{examples:tgre_rank_skill_1}
\end{figure*}

\begin{figure*}[!ht]
\centering
\begin{tcolorbox}[colback=blue!5!white, colframe=blue!75!black, title=Example 2 for TGRE-based Reranking Prompt]
\small
\begin{VerbatimWrap}
Text snippet: You are proficient in Python and English
Options:
APL; Ancient Greek; Arabic; C++; COBOL; English; Erlang; French; German; Greek; Groovy; ICT system programming; Japanese; Java (computer programming); Latin; MATLAB; ML (computer programming); Objective-C; Pascal (computer programming); Python (computer programming); Ruby (computer programming); Russian; Scala; TypeScript; Yiddish; academic English; aviation English; comprehend the material to be translated; computer programming; computer science; computer technology; develop digital educational materials; e-learning; interpret languages in conferences; keywords in digital content; manage business knowledge; manage localisation; query languages; teach ESOL language class; teach ESOL literacy class; teach languages; teach linguistics; technical terminology; technology education; translate foreign language; translate spoken language; understand spoken English; understand written English; use localisation tools; write English
Reasoning: The job vacancy snippet explicitly mentions "Python" and "English." These phrases give us clues in determining the most relevant options by aligning them with the specific requirements in the snippet.
- The term "Python" directly corresponds to [Python (computer programming): \textcolor{blue}{The techniques and principles of software development, such as analysis, algorithms, coding, testing and compiling of programming paradigms in Python}], making it the most relevant option.
- The term "English" directly corresponds to [English: \textcolor{blue}{The English language}.] and it can also be relevant to "academic English," "understand spoken English," "understand written English," and "write English," depending on the context in which English proficiency is required.
Therefore, we select the following 10 most applicable options:
1. Python (computer programming)
2. English
3. understand written English
4. understand spoken English
5. write English
6. academic English
7. ICT system programming
8. computer programming
9. technical terminology
10. technology education
\end{VerbatimWrap}
\end{tcolorbox}
\caption{In-context example 2 for TGRE-based reranking prompts for skill classification. Segments highlighted in blue are detailed description of the corresponding occupation retrieved from the taxonomy.}
\label{examples:tgre_rank_skill_2}
\end{figure*}

\begin{figure}[!ht]
\centering
\begin{tcolorbox}[colback=blue!5!white, colframe=blue!75!black, title=Example 1 for CoT-based Inference Prompt]
\small
\begin{VerbatimWrap}
Vacancy: performed code reviews for your teammates
Reasoning: Let's think step by step in order to find the skills. We need to find the skills that are mentioned in the vacancy.
Skills: Code review, teamwork
\end{VerbatimWrap}
\end{tcolorbox}
\begin{tcolorbox}[colback=blue!5!white, colframe=blue!75!black, title=Example 2 for CoT-based Inference Prompt]
\small
\begin{VerbatimWrap}
Vacancy: You are a good communicator and enjoy working in a team
Reasoning: Let's think step by step in order to find the skills. We need to find the skills that are mentioned in the vacancy.
Skills: Communication, Teamwork
\end{VerbatimWrap}
\end{tcolorbox}
\caption{In-context examples for CoT-based inference prompts GPT-3.5 Turbo for skill classification}
\label{examples:cot_infer_skill}
\end{figure}

\begin{figure*}[!ht]
\centering
\begin{tcolorbox}[colback=blue!5!white, colframe=blue!75!black, title=Example 1 for CoT-based Reranking Prompt]
\small
\begin{VerbatimWrap}
Text snippet: performed code reviews for your teammates
Options: 
work in fitness teams; collaborate in company's daily operations; risk management; work in assembly line teams; lead a team; set up effective working relationships with other sports players; craftsmanship; encourage teambuilding; capacity building; work as a team in a hazardous environment; peer group methods; team building; build team spirit; supervise crew; motivate others; organise supporters; plan the work of teams and individuals; social alliances; teamwork principles; conduct ICT code review; military drill; lead a team in forestry services; work in teams; stimulate creativity in the team; manage the security team; coordinate engineering teams; plan teamwork; coaching techniques; manage a team; perform inspection analysis; assemble an artistic team; encourage teams for continuous improvement; personnel management; work in a construction team; motivate supporters; collaborate with coaching team; create social alliances; establish collaborative relations; coach staff for running the performance; evaluate performance of organisational collaborators; communication; report to the team leader; coordinate communication within a team; work with supporting team in community arts programme; contribute to the programmer's reflection process; cooperate with colleagues; ensure positive relations within the artistic team; write inspection reports; facilitate teamwork between students; shape organisational teams based on competencies
Reasoning: Let's think step by step in order to identify the most applicable skills from the snippet. When the snippet mentions "performed code reviews for your teammates," it highlights the need for skills related to understanding and assessing code quality, providing constructive feedback, and collaborating effectively within a technical team context.
Skills: conduct ICT code review; teamwork principles; communication; facilitate teamwork between students; coordinate communication within a team; evaluate performance of organisational collaborators
\end{VerbatimWrap}
\end{tcolorbox}
\caption{In-context example 1 for CoT-based prompts generated by GPT-3.5 Turbo for skill classification}
\label{examples:cot_rank_skill_1}
\end{figure*}

\begin{figure*}[!ht]
\centering
\begin{tcolorbox}[colback=blue!5!white, colframe=blue!75!black, title=Example 2 for CoT-based Reranking Prompt]
\small
\begin{VerbatimWrap}
Text snippet: You are a good communicator and enjoy working in a team
Options:
use communication devices; inter-professional communication in the psychosocial field; encourage teambuilding; stimulate creativity in the team; disseminate internal communications; coordinate communication within a team; lead a team; use therapeutic communication techniques; communication; team building; promote good habits to avoid communication disorders; disseminate messages to people; communication related to hearing impairment; teamwork principles; communicate with media; work in teams; manage a team; digital communication and collaboration; communication studies; maintain internal communication systems; handle communications in the food processing industry; plan teamwork; body language; circulate information; cognitive psychology; plan the work of teams and individuals; cooperate with colleagues; teach communication methods; encourage teams for continuous improvement; communication disorders; communicate by use of interpretation services; communication principles; use communication equipment; use communication techniques; communication sciences; facilitate teamwork between students; teach communication to clients; apply technical communication skills; communicate by telephone; advise on communication strategies; promote organisational communication; use different communication channels; electronic communication; use a complex communication system; communicate in healthcare; promote communication between parties; develop communications strategies; develop inclusive communication material; psychology; build team spirit
Reasoning: Let's think step by step in order to identify the most applicable skills from the snippet. We need to focus on skills directly related to communication and teamwork. The snippet mentions being a "good communicator" and "enjoy working in a team," so we will prioritize skills that align with these two aspects.
Skills: communication; teamwork principles; work in teams; use communication techniques; promote communication between parties; coordinate communication within a team; build team spirit; cooperate with colleagues; lead a team; manage a team
\end{VerbatimWrap}
\end{tcolorbox}
\caption{In-context example 2 for CoT-based prompts generated by GPT-3.5 Turbo for skill classification}
\label{examples:cot_rank_skill_2}
\end{figure*}

\section{Supplemental Tables}
\label{sec:supp_tables}

This section provides supplemental tables supporting the results discussed in the main paper. Table~\ref{tab:lc_score} reports the Precision@1 scores for Lightcast's default SOC labels in the occupation classification task, evaluated by GPT-4o. Table~\ref{tab:finetuning_score} summarizes RP@5 and RP@10 results for two fine-tuned systems~\cite{decorte2023extreme,clavie2023large} in the skill classification evaluation. Additionally, Tables~\ref{tab:skills_results_3} and~\ref{tab:skills_results_5} present the RP@3 and RP@5 results for skill classification.

\begin{table}[!th]
\centering
\caption{Precision@1 for Lightcast's default SOC labels in occupation classification evaluation.}
\label{tab:lc_score}
\begin{tabular}{@{}ll@{}}
\toprule
Method & Precision@1 \\ \midrule
LC & 0.634 \\ \bottomrule
\end{tabular}
\end{table}

\begin{table}[!ht]
\centering
\caption{RP@5 and RP@10 for two baseline fine-tuned systems in skill classification evaluation.}
\label{tab:finetuning_score}
\begin{tabular}{@{}lll@{}}
\toprule
Method & RP@5 & RP@10 \\ \midrule
Decorte et al. & 0.5462 & N/A \\
Clavi{\'e} et al. & 0.615 & 0.6894 \\ \bottomrule
\end{tabular}
\end{table}

\begin{table*}[!h]
\caption{RP@3 for skill classification. The best results are highlighted in bold. The columns abbreviate the full names of the LLMs: GPT refers to GPT-3.5 Turbo, Gemini to Gemini-1.5 Flash, Claude to Claude Instant 1.2, Llama to Llama 3 8B Instruct, and Mistral to Mistral Small. For TGRE* and CoT*, we use GPT-4o for reranking.}
\label{tab:skills_results_3}
\centering
\begin{tabular}{@{}lllrrrrr@{}}
\toprule
Baseline & Reasoning & Retrieval & \multicolumn{1}{l}{GPT} & \multicolumn{1}{l}{Gemini} & \multicolumn{1}{l}{Claude} & \multicolumn{1}{l}{Llama} & \multicolumn{1}{l}{Mistral} \\ 
Equivalence & Method & Unit &  & & & & \\ \midrule
& TGRE* & Sentence & \textbf{0.5237} & \textbf{0.5685} & \textbf{0.5419} & \textbf{0.5311} & \textbf{0.5444} \\
& TGRE & Sentence & 0.4423 & 0.4689 & 0.4428 & 0.4053 & 0.4561 \\
& CoT & Sentence & 0.177 & 0.2318 & 0.3018 & 0.2002 & 0.2229 \\
\midrule
& TGRE & Label & 0.4034 & 0.4532 & 0.4181 & 0.3491 & 0.4186 \\
IReRa & CoT* & Label & 0.4132 & 0.4758 & 0.4339 & 0.4231 & 0.4502 \\
IReRa & CoT & Label & 0.4778 & 0.4053 & 0.3841 & 0.2816 & 0.3851 \\
\midrule
LLM4Jobs & TGRE- & Sentence & 0.4201 & 0.4389 & 0.4398 & 0.4438 & 0.4142 \\
LLM4Jibs & CoT- & Sentence & 0.0868 & 0.0863 & 0.1677 & 0.0932 & 0.1179 \\
& TGRE- & Label & 0.4334 & 0.502 & 0.4679 & 0.3496 & 0.4127 \\
& CoT- & Label & 0.4157 & 0.3683 & 0.2411 & 0.2396 & 0.2791 \\
& TGRE-\,- & None & 0.1543 & 0.1045 & 0.1134 & 0.0996 & 0.1011 \\
& CoT-\,- & None & 0.0197 & 0.0316 & 0.0079 & 0.0015 & 0.0266 \\ \bottomrule
\end{tabular}
\end{table*}

\begin{table*}[!h]
\caption{RP@5 for skill classification. The best results are highlighted in bold. The columns abbreviate the full names of the LLMs: GPT refers to GPT-3.5 Turbo, Gemini to Gemini-1.5 Flash, Claude to Claude Instant 1.2, Llama to Llama 3 8B Instruct, and Mistral to Mistral Small. For TGRE* and CoT*, we use GPT-4o for reranking.}
\label{tab:skills_results_5}
\centering
\begin{tabular}{@{}lllrrrrr@{}}
\toprule
Baseline & Reasoning & Retrieval & \multicolumn{1}{l}{GPT} & \multicolumn{1}{l}{Gemini} & \multicolumn{1}{l}{Claude} & \multicolumn{1}{l}{Llama} & \multicolumn{1}{l}{Mistral} \\ 
Equivalence & Method & Unit &  & & & & \\ \midrule
& TGRE* & Sentence & \textbf{0.6075} & \textbf{0.6318} & \textbf{0.643} & \textbf{0.6107} & \textbf{0.6004} \\
& TGRE & Sentence & 0.5252 & 0.5681 & 0.5258 & 0.4681 & 0.5226 \\
& CoT & Sentence & 0.1915 & 0.2677 & 0.352 & 0.2359 & 0.2587 \\
\midrule
& TGRE & Label & 0.463 & 0.5338 & 0.4974 & 0.3989 & 0.4747 \\
IReRa & CoT* & Label & 0.4851 & 0.5675 & 0.524 & 0.4857 & 0.5323 \\
IReRa & CoT & Label & 0.5244 & 0.4832 & 0.4529 & 0.334 & 0.4407 \\
\midrule
LLM4Jobs & TGRE- & Sentence & 0.5078 & 0.5163 & 0.5065 & 0.5033 & 0.4901 \\
LLM4Jobs & CoT- & Sentence & 0.1223 & 0.1072 & 0.2071 & 0.1168 & 0.1473 \\
& TGRE- & Label & 0.4857 & 0.5469 & 0.5131 & 0.4123 & 0.4704 \\
& CoT- & Label & 0.4929 & 0.4381 & 0.3213 & 0.2843 & 0.3484 \\
& TGRE-\,- & None & 0.1523 & 0.1045 & 0.113 & 0.0994 & 0.1009 \\
& CoT-\,- & None & 0.0301 & 0.0459 & 0.0136 & 0.0054 & 0.039 \\ \bottomrule
\end{tabular}
\end{table*}

\section{Supplemental Figures}
\label{sec:supp_figures}

This section presents supplemental figures supporting the results discussed in the main body of the paper.

\begin{figure*}[ht]
    \centering
    \includegraphics[width=0.9\textwidth]{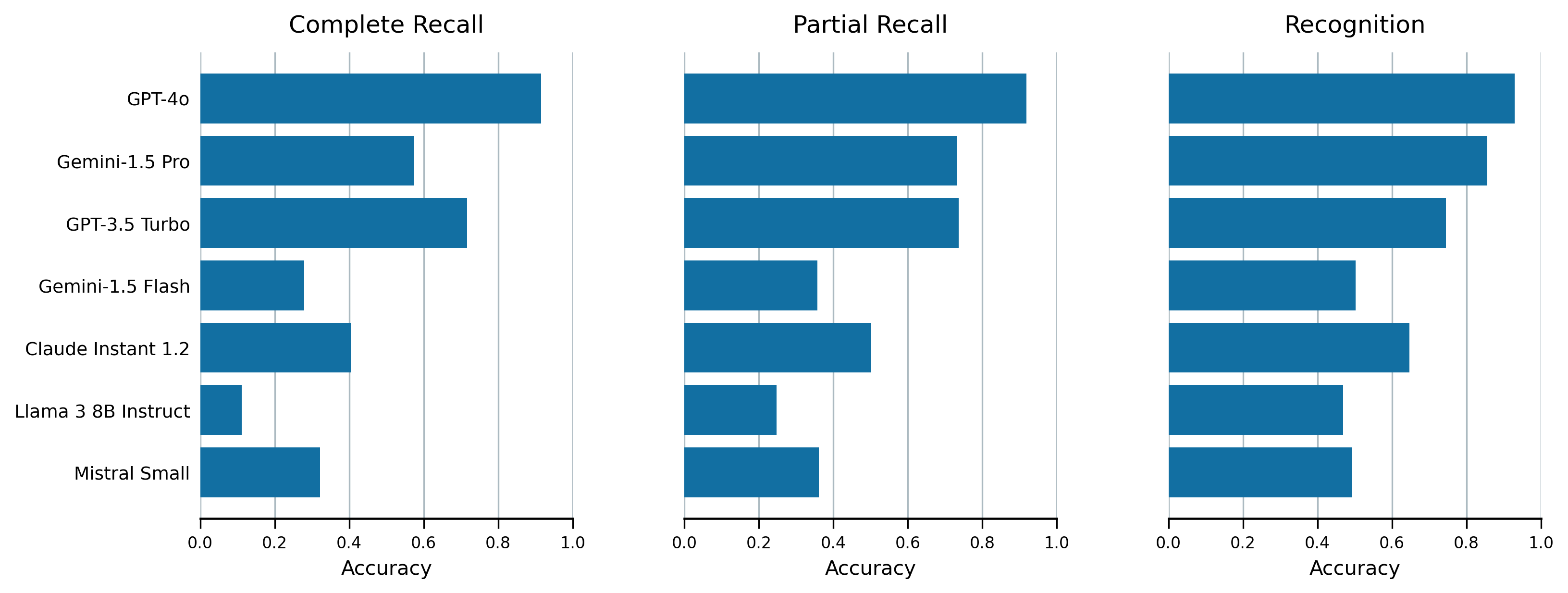}
    \caption{The performance of different LLMs on the complete recall, partial recall, and recognition tasks.}
    \label{fig:knowledge_models_recall}
\end{figure*}

\begin{figure*}[ht]
    \centering
    \includegraphics[width=0.9\textwidth]{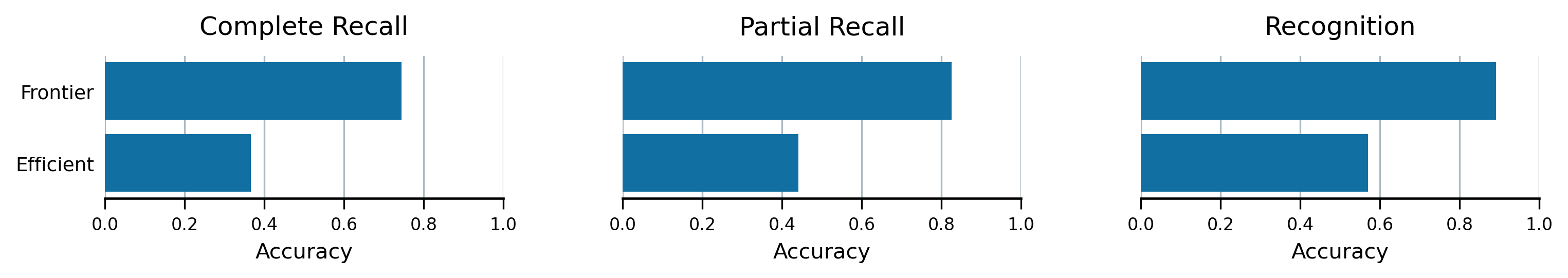}
    \caption{The performance of frontier and efficient LLMs on the complete recall, partial recall, and recognition tasks.}
    \label{fig:knowledge_model_groups_recall}
\end{figure*}

\begin{figure*}[ht]
    \centering
    \includegraphics[width=0.9\textwidth]{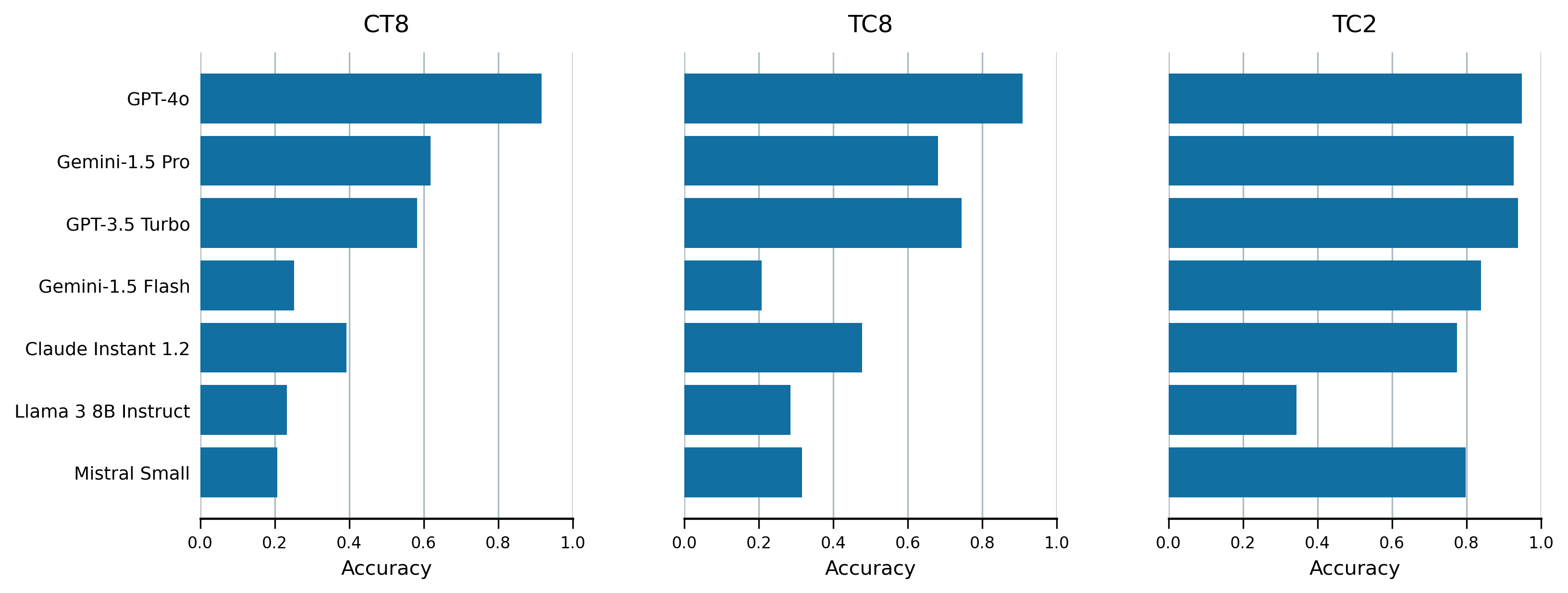}
    \caption{The performance of different LLMs on tasks at 8-digit (CT8 and TC8) and 2-digit codes (TC2).}
    \label{fig:knowledge_models_granularity}
\end{figure*}

\begin{figure*}[ht]
    \centering
    \includegraphics[width=0.9\textwidth]{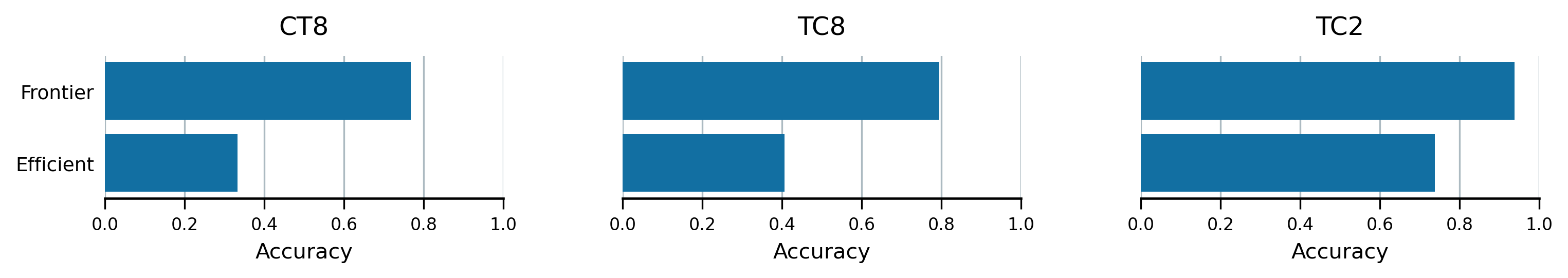}
    \caption{The performance of frontier and efficient LLMs on tasks at 8-digit (CT8 and TC8) and 2-digit codes (TC2).}
    \label{fig:knowledge_model_groups_granularity}
\end{figure*}

Figures~\ref{fig:knowledge_models_recall} and \ref{fig:knowledge_model_groups_recall} illustrate the accuracy scores for the complete recall, partial recall, and recognition tasks of individual LLMs as well as frontier and efficient LLMs. Figures~\ref{fig:knowledge_models_granularity} and \ref{fig:knowledge_model_groups_granularity} show the accuracy scores for tasks involving 8-digit and 2-digit codes for both individual LLMs and frontier/efficient LLMs. As highlighted in Section~\ref{subsec:assessing_results}, GPT-4o emerges as the best overall performer, followed by GPT-3.5 Turbo and Gemini-1.5 Pro. Furthermore, frontier models consistently outperform efficient models across all tasks.

\begin{figure}[ht]
    \centering
    \includegraphics[width=\columnwidth]{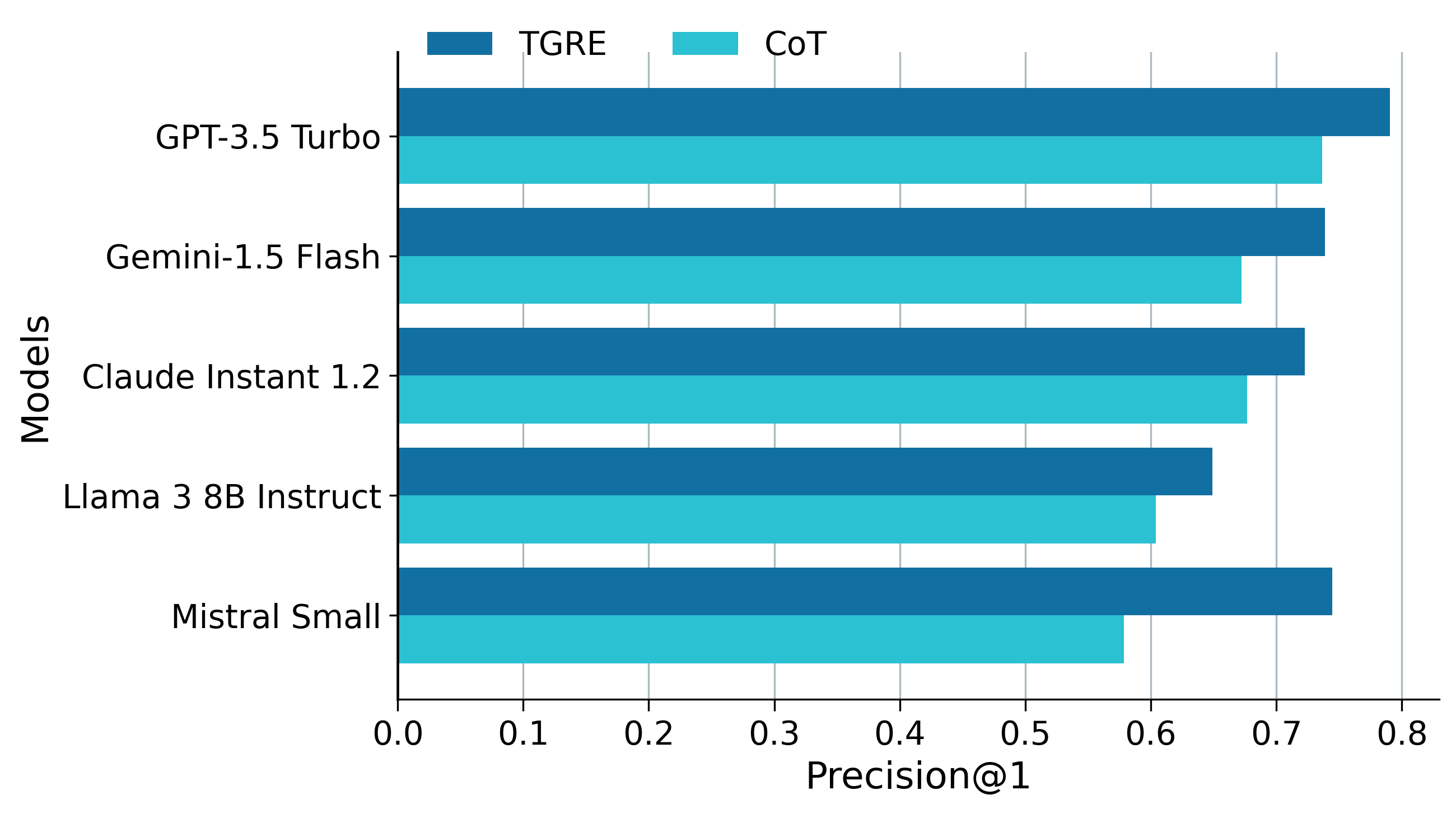}
    \caption{The performance of different reasoning methods on occupation classification.}
    \label{fig:onet_reasonings}
\end{figure}

\begin{figure}[ht]
    \centering
    \includegraphics[width=\columnwidth]{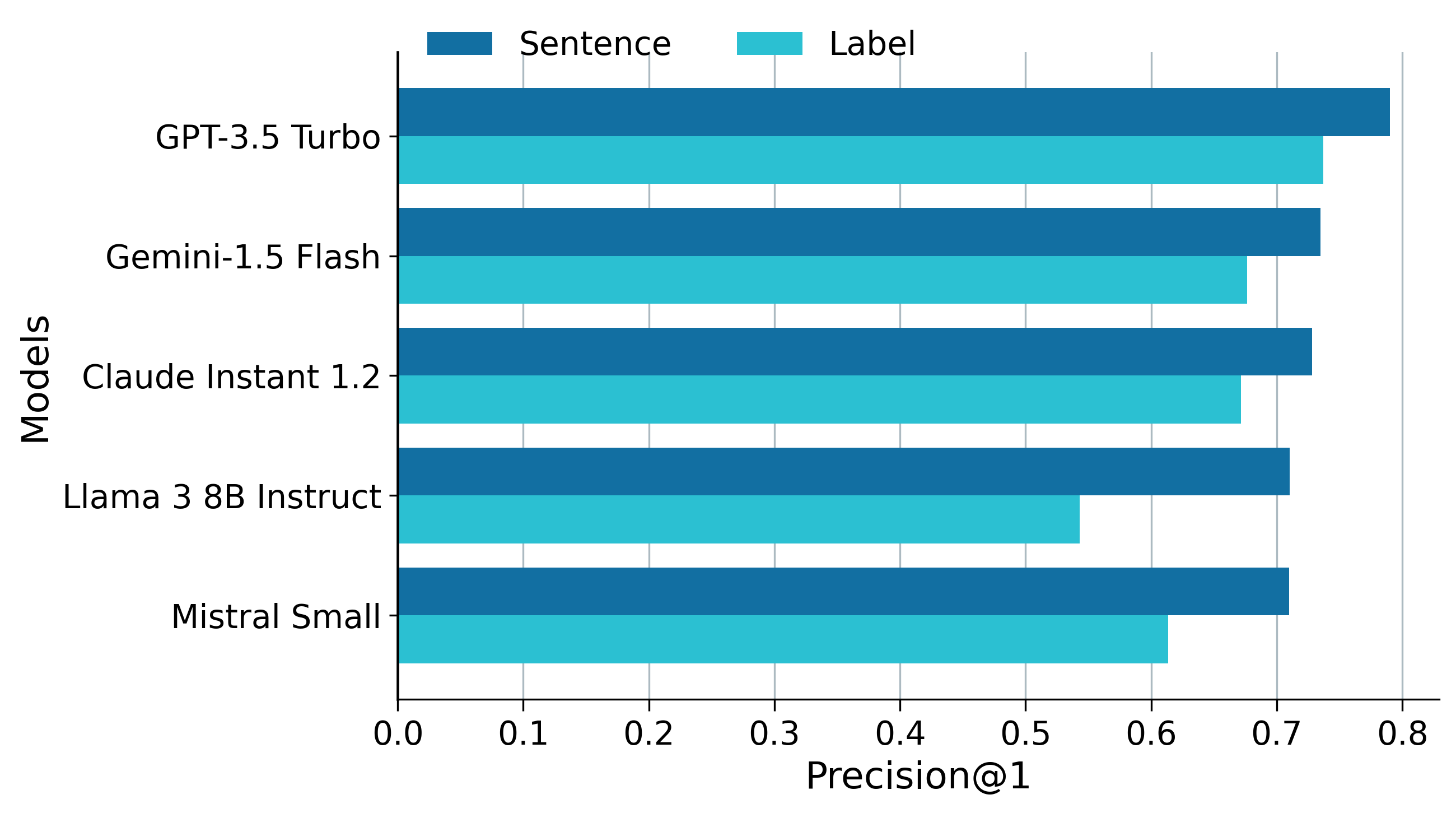}
    \caption{The performance of different retrieval units on occupation classification.}
    \label{fig:onet_retrieval_units}
\end{figure}

Figures~\ref{fig:onet_reasonings} and \ref{fig:onet_retrieval_units} depict Precision@1 results for various reasoning methods and retrieval units in occupation classification, respectively. As discussed in Section~\ref{subsec:occupation_results}, TGRE-based methods demonstrate superior performance compared to CoT-based methods, while sentence-based retrieval units outperform label-based retrieval units.

\begin{figure}[ht]
    \centering
    \includegraphics[width=\columnwidth]{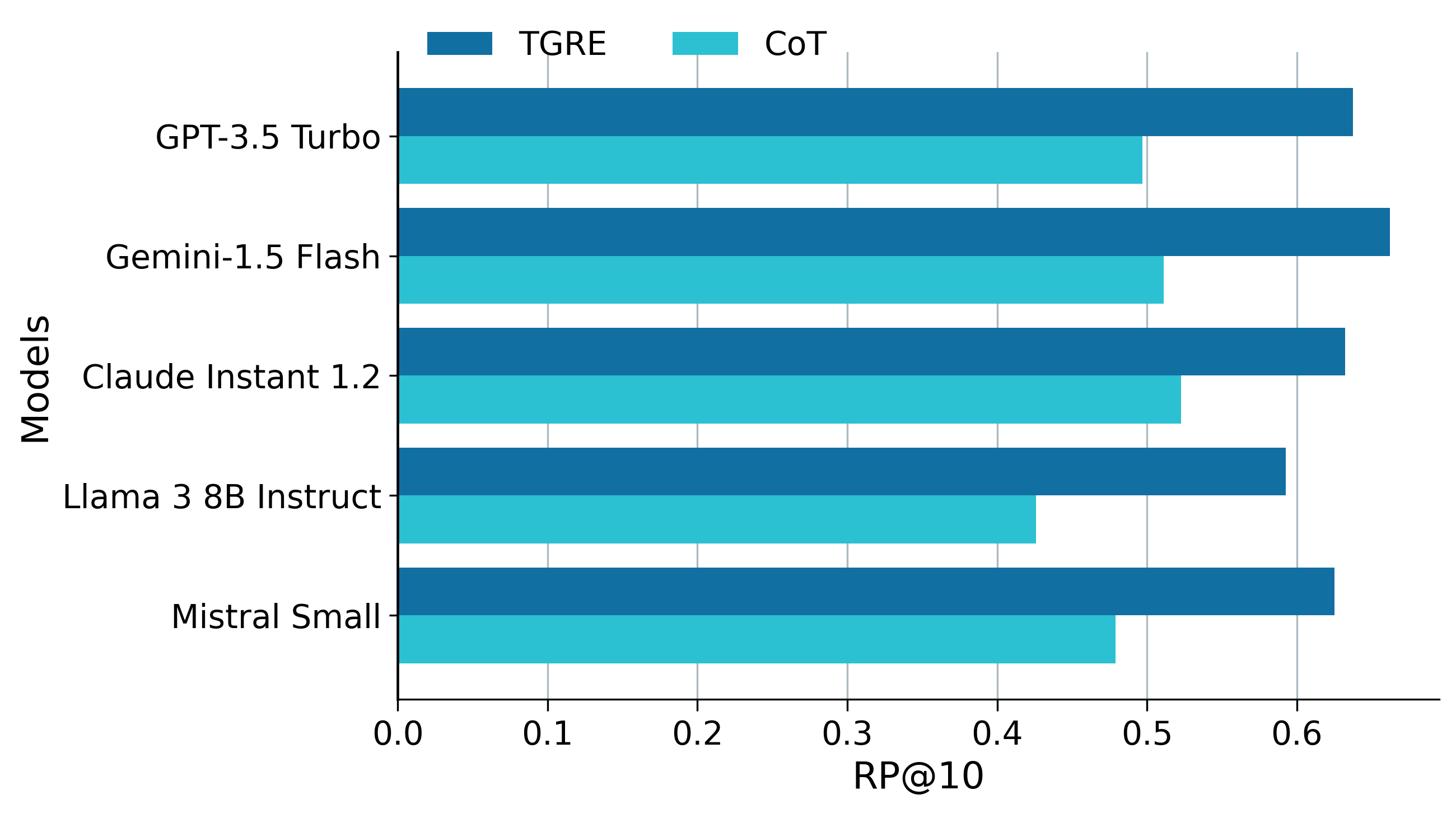}
    \caption{The performance of different reasoning methods on skill classification.}
    \label{fig:esco_reasonings}
\end{figure}

\begin{figure}[ht]
    \centering
    \includegraphics[width=\columnwidth]{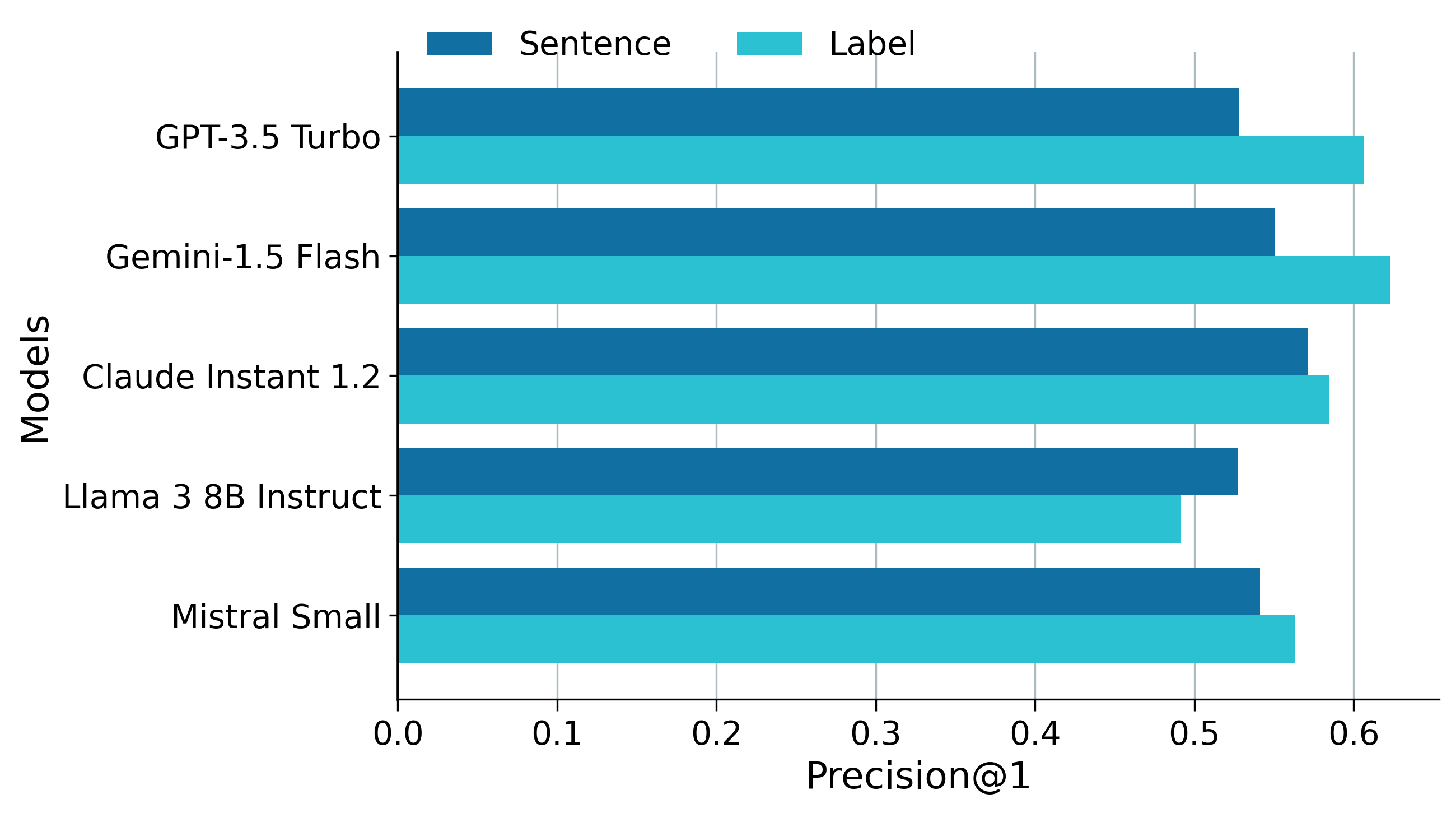}
    \caption{The performance of different retrieval units on skill classification.}
    \label{fig:esco_retrieval_units}
\end{figure}

Likewise, Figures~\ref{fig:esco_reasonings} and \ref{fig:esco_retrieval_units} present RP@10 results for various reasoning methods and retrieval units in skill classification. As highlighted in Section~\ref{subsec:skill_results}, TGRE-based methods achieve better performance than CoT-based methods. However, label-based retrieval units exhibit a higher average RP@10 compared to sentence-based retrieval units in most LLMs, influenced by some outliers (CoT with sentence-based retrieval variants). Nonetheless, a one-way ANOVA with post-hoc analysis indicates that sentence-based retrieval units significantly outperform label-based retrieval units ($p<0.05$).

\section{Computational Cost Analysis}
\label{sec:cost_analysis}

\subsection{Occupation Classification}
\label{subsec:cost_analysis_occupation}

We conducted a study using 100 randomly sampled records (job title-company name pairs) from the test set to assess the computational costs of TGRE compared to GPT-4o's direct prompting. GPT-3.5 Turbo was used as an underlying LLM for TGRE, whereas GPT-4o directly inferred SOC labels using CoT reasoning with the same CoT-based inference template (Figure~\ref{prompt:cot_occupation}). Both models were accessed via OpenAI APIs on March 18, 2025. Each test record was processed through a single API call, with no concurrent calls. The study was performed on a commodity machine equipped with an Intel Core i9-13900H 2600 MHz CPU (16 GB RAM) and an NVIDIA GeForce RTX 4060 GPU (16 GB VRAM).

We measured the following for each record to provide a comprehensive view of computational cost-effectiveness.

\begin{itemize}
    \item \textbf{Average Tokens:} Computed as the total token usage (input tokens in the input context + output tokens generated by the model) summed across all records and divided by 100.
    \item \textbf{Average Cost ($\times 10^{-3}$ USD):} Computed as the sum of costs per record, based on the API token pricing. As of March 2025\footnote{https://openai.com/api/pricing/}, GPT-3.5 Turbo costs \$0.5 per 1M input tokens and \$1.5 per 1M output tokens, while GPT-4o costs \$2.5 per 1M input tokens and \$10 per 1M output tokens.
    \item \textbf{Average Execution Time (seconds):} Computed as the sum of API call latency (in seconds) for all records and divided by 100. For TGRE, this metric also includes embedding generation (0.0966 seconds) and retrieval time (0.0307 seconds).
    \item \textbf{Throughput (records per second):} Computed as the inverse of execution time.
\end{itemize}

\begin{table}[th!]
\centering
\resizebox{\columnwidth}{!}{%
\begin{tabular}{@{}lccc@{}}
\toprule
\multirow{2}{*}{\textbf{Metric}} & \multirow{2}{*}{\textbf{TGRE}} & \multirow{2}{*}{\textbf{GPT-4o}} & \textbf{TGRE} \\ 
& & & \textbf{(Inference Only)} \\ \midrule
Tokens & 986.5 $\pm$ 5.7 & 570.8 $\pm$ 15.4 & 262.8 $\pm$ 11.7 \\
Cost & 0.78 $\pm$ 0.04 & 2.76 $\pm$ 0.15 & 0.2 $\pm$ 0.02 \\
Time & 4.01 $\pm$ 0.57 & 2.88 $\pm$ 0.7 & 1.34 $\pm$ 0.18 \\ \bottomrule
\end{tabular}}
\caption{Computational cost comparison across models for occupation classification. The values represent the mean $\pm$ standard deviation.}
\label{tab:cost_analysis}
\end{table}

As shown in Table~\ref{tab:cost_analysis}, TGRE processes 72.83\% more tokens on average than GPT-4o. This is logical as TGRE involves both the inference and reranking stages, whereas GPT-4o only processes tokens during the inference stage. However, when comparing only the inference stages, TGRE is significantly more token-efficient, processing 53.96\% fewer tokens than GPT-4o.

TGRE also achieves a substantial cost reduction of 71.95\% compared to GPT-4o. This demonstrates the cost-effectiveness when applied to smaller, more efficient LLMs. For large-scale classification tasks involving one million input instances, the total costs of TGRE with GPT-3.5 Turbo and GPT-4o amount to \$774.72 and \$2,761.78, respectively.

As for execution time, TGRE takes approximately one second longer per record (39.33\%), compared to GPT-4o, due to the reranking step. Nevertheless, it remains highly efficient when comparing the execution time of their inference stages alone. Based on the average execution time, the estimated throughputs for TGRE and GPT-4o are approximately 0.26 records per second and 0.35 records per second, respectively. Additionally, TGRE incurs a one-time overhead for generating embeddings for O*NET-SOC, which took about 455.65 seconds for all 1,016 SOC entities, averaging 0.45 seconds per entity.

In conclusion, The analysis shows that TGRE maintains a balance between computation cost and classification accuracy. While GPT-4o's direct prompting is more streamlined and marginally faster, TGRE's remarkable cost advantage makes it a practical and economical choice for large-scale classifications. The inference-only comparison further emphasizes its token efficiency and processing speed.

\subsection{Skill Classification}
\label{subsec:cost_analysis_skill}

We applied the same methodology to analyze computational costs for skill classification as was used for occupation classification. As shown in Table~\ref{tab:cost_analysis_skill}, TGRE's token usage for skill classification is significantly higher compared to occupation classification tasks. This discrepancy is due to the inherent differences in task nature. Unlike the multi-class occupation classification, skill classification is a multi-label classification problem, requiring the inference and reranking of multiple labels per test instance. This results in higher token consumption and generation throughout the framework.

\begin{table}[th!]
\centering
\resizebox{\columnwidth}{!}{%
\begin{tabular}{@{}lccc@{}}
\toprule
\multirow{2}{*}{\textbf{Metric}} & \multirow{2}{*}{\textbf{TGRE}} & \multirow{2}{*}{\textbf{GPT-4o}} & \textbf{TGRE} \\ 
& & & \textbf{(Inference Only)} \\ \midrule
Tokens & 2110.13 $\pm$ 31.06 & 288 $\pm$ 34.75 & 615.4 $\pm$ 56.09 \\
Cost & 1.34 $\pm$ 0.09 & 1.45 $\pm$0.33 & 0.46 $\pm$ 0.08 \\
Time & 4.12 $\pm$ 0.96 & 1.78 $\pm$ 0.62 & 1.97 $\pm$ 0.7 \\ \bottomrule
\end{tabular}}
\caption{Computational cost comparison across models fro skill classification. The values represent the mean $\pm$ standard deviation.}
\label{tab:cost_analysis_skill}
\end{table}

Interestingly, GPT-4o's direct prompting exhibits greater token efficiency in skill classification compared to its performance in occupation classification tasks. Despite this, TGRE still achieves an approximate 7.45\% cost reduction relative to GPT-4o, though the gap is considerably smaller than in the occupation classification scenario. 

More importantly, the inclusion of GPT-4o's cost comparison is mainly for completeness. Unlike occupation classification, GPT-4o's direct prompting approach in skill classification is severely hindered by a few major limitations. Firstly, GPT-4o exhibits a significantly limited knowledge of the ESCO taxonomy compared to ONET-SOC. A preliminary knowledge assessment study reveals GPT-4o's accuracy in generating complete ESCO occupations to be only 0.010 -- a stark contrast to its 0.9211 accuracy for ONET-SOC. This disparity strongly suggests that GPT-4o's capability in ESCO skill inference is similarly limited. Secondly, When evaluated on the 100 random test instances used in the computational cost analysis, GPT-4o's direct prompting yields an RP@10 score of only 0.015. This demonstrates its inability to effectively predict ESCO skills without leveraging our framework.

\end{document}